\documentclass[AMS,STIX2COL,SYSTEMFONTS]{WileyNJD-v2}

\usepackage{amsmath,amsfonts}
\usepackage{adjustbox}
\usepackage{color,soul}
\usepackage{anyfontsize}
\articletype{Research Article}%

\received{26 April 2016}
\revised{6 June 2016}
\accepted{6 June 2016}

\begin{document}

\title{Nonparametric clustering for image segmentation}

\author[1]{Giovanna Menardi*}

\authormark{G. Menardi}

\address[1]{\orgdiv{Department of Statistical Sciences}, \orgname{University of Padova}, \orgaddress{\state{Padova}, \country{Italy}}}

\corres{*Giovanna Menardi, \email{menardi@stat.unipd.it}}

\presentaddress{via C. Battisti 241, 35121 Padova, Italy}

\abstract[Summary]{Image segmentation aims at identifying regions of interest within an image, by grouping
pixels according to their properties. This task resembles the statistical one of clustering, yet 
many standard clustering methods fail to meet the basic requirements of image segmentation: segment shapes
are often biased toward predetermined shapes and their number is rarely determined automatically.
Nonparametric clustering is, in principle, free from these limitations and turns out to be particularly suitable for the task of image segmentation.
This is also witnessed by several operational analogies, as, for instance, the resort to topological data analysis and
spatial tessellation in both the frameworks.

We discuss the application of nonparametric clustering to image segmentation
and provide an algorithm specific for this task. Pixel similarity is evaluated in terms of density of the color representation and the
adjacency structure of the pixels is exploited to introduce a simple, yet effective method to
identify image segments as disconnected high-density regions. The proposed method works both to segment an image and to
detect its boundaries and can be seen as a generalization to color images of the class of thresholding methods.}


\jnlcitation{\cname{%
A revised version of this manuscript is the following:\\
\author{Menardi, G.}, 
(\cyear{2020}), 
\ctitle{Nonparametric clustering for image segmentation}, \cjournal{Statistical Analysis and DataMining: The ASA Data Science Journal}, \cvol{13}(1), p. 83-97.}}

\maketitle


\section{Introduction and motivation}\label{sec:intro}

In the recent years, the need of analyzing large amounts of image information has become relevant in several contexts
and applications. Daily examples include medical diagnosis based on X-ray or magnetic resonance images, video surveillance
and geographic information system applications, and image tagging. 
A possible goal of image analysis is the one of \emph{segmentation}, the automatic process of
identifying salient regions and single objects in an image, with the purpose of content retrieval, object detection
or recognition, occlusion boundary, image compression or editing. 

Digital images are created by a variety of input devices, such as cameras or scanners,
and they have usually a fixed resolution, \emph{i.e.} they are represented by a fixed number of digital values,
known as \emph{pixels}. Pixels are the smallest individual element in an image, holding quantized values
that represent the brightness of a given color at any specific location of the image.
When an image is segmented, a label is assigned to each pixel, so that pixels with the same
label share similar characteristics in terms of color, intensity, or texture.

This task recalls closely the aim of cluster analysis, and thereby clustering methods
have been featured as a standard tool to segment images. Within this framework, 
an approach which naturally lends itself to the task of image segmentation 
is known as \emph{nonparametric} or \emph{modal} clustering. 
{With respect to most of clustering methods, relying on heuristic ideas of similarity between objects,  
the nonparametric formulation claims a sounder theoretical ground, since the definition of a precise notion of clusters 
allows for a ``ground truth'' to aim at when evaluating a clustering configuration or comparing alternatives. }
Specifically, 
a probability density function is assumed to underlie the data and 
clusters are defined as the domains of attraction of the modes
of the density function, estimated nonparametrically. 

The correspondence between clusters and regions around the modes of the data
distribution entails some reasons of attractiveness.
First, the number of clusters is an intrinsic
property of the data generator mechanism, thereby conceptually well defined, and its determination is itself an integral part of the clustering procedure.
Additionally, modal regions comply with the geometric intuition
about the notion of clusters, also because they are not bound to any particular shape.
These reasons make nonparametric clustering particularly suitable for the segmentation of digital images,
as segments shall be allowed to assume arbitrary shapes and an automatic determination of the number of segments
would be desirable.  

In this work the use of nonparametric clustering for image segmentation is discussed.
Pixel similarity is evaluated in terms of density of the color representation and the
adjacency structure of the pixels is exploited to introduce a simple method to
assess the connectedness of the modal density regions. 

In the following, an overview about nonparametric clustering is provided along with its connection
with methods for image segmentation. Within this framework, a novel method specifically conceived for
image segmentation is proposed and discussed, and several applications illustrated. 
  
\section{Background}
\subsection{Overview of nonparametric clustering}
\begin{figure*}[t]
\begin{center}
\begin{tabular}{ccc}
\hspace{-.5cm}\includegraphics[width=.3\textwidth]{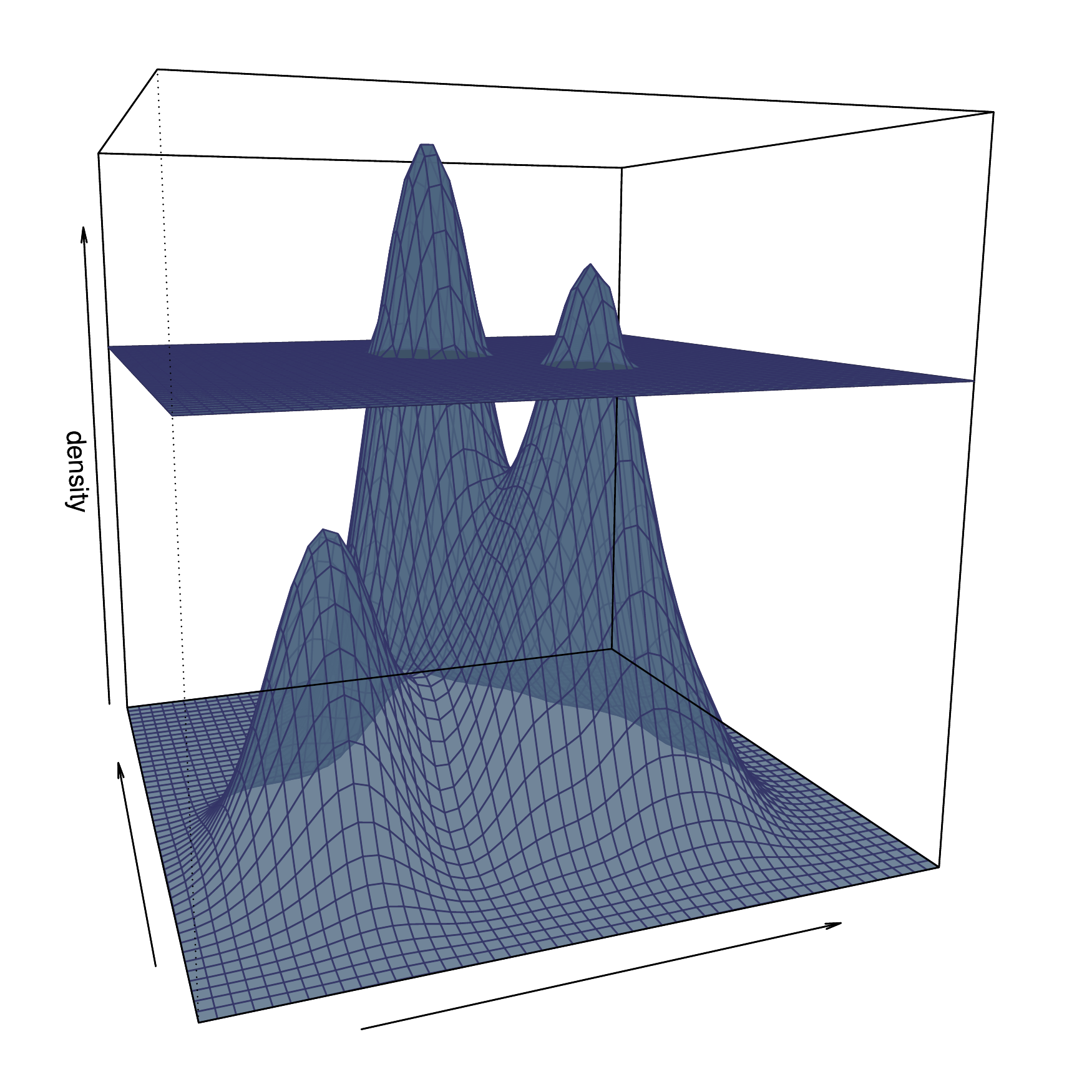} &
\includegraphics[width=.3\textwidth]{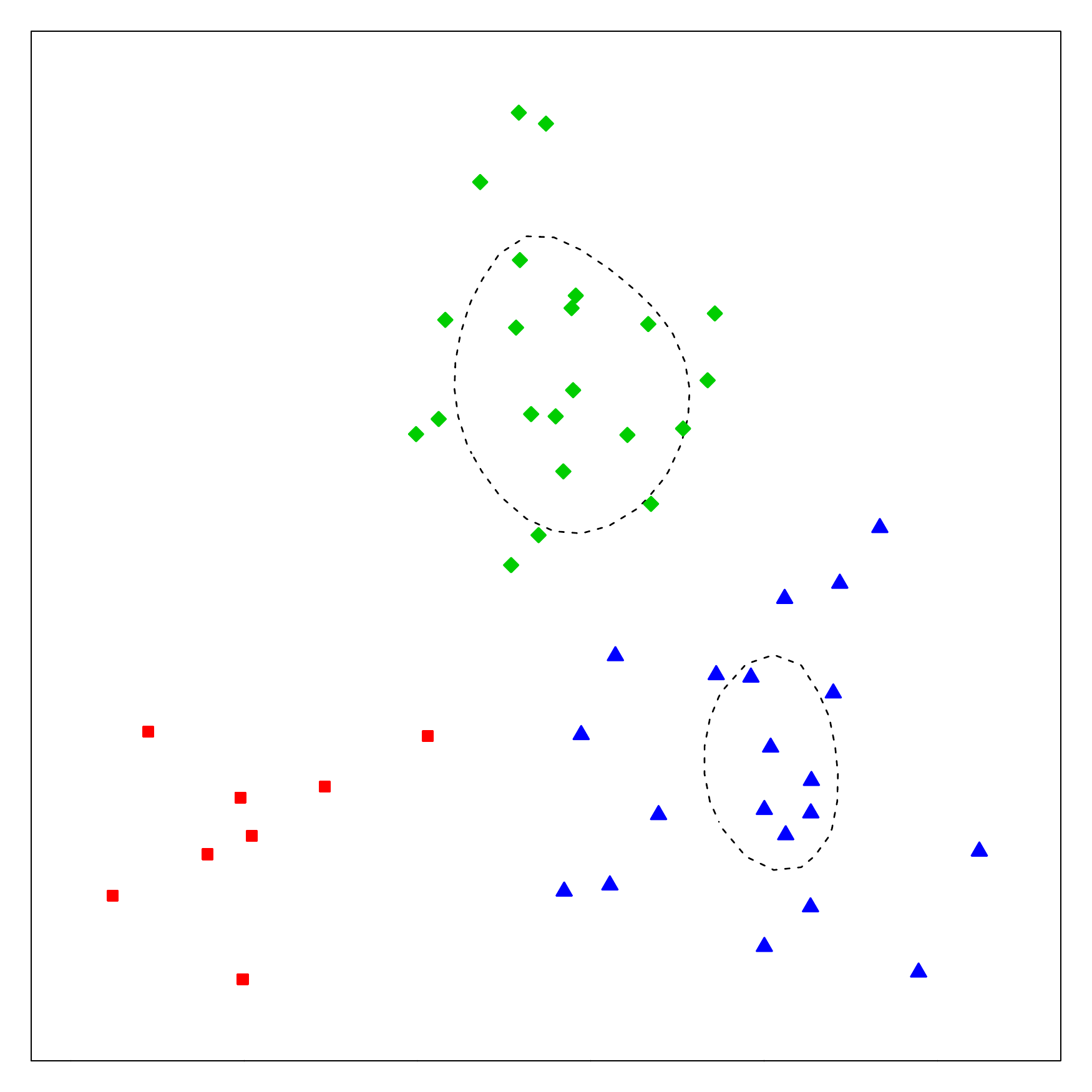}&
\includegraphics[width=.3\textwidth]{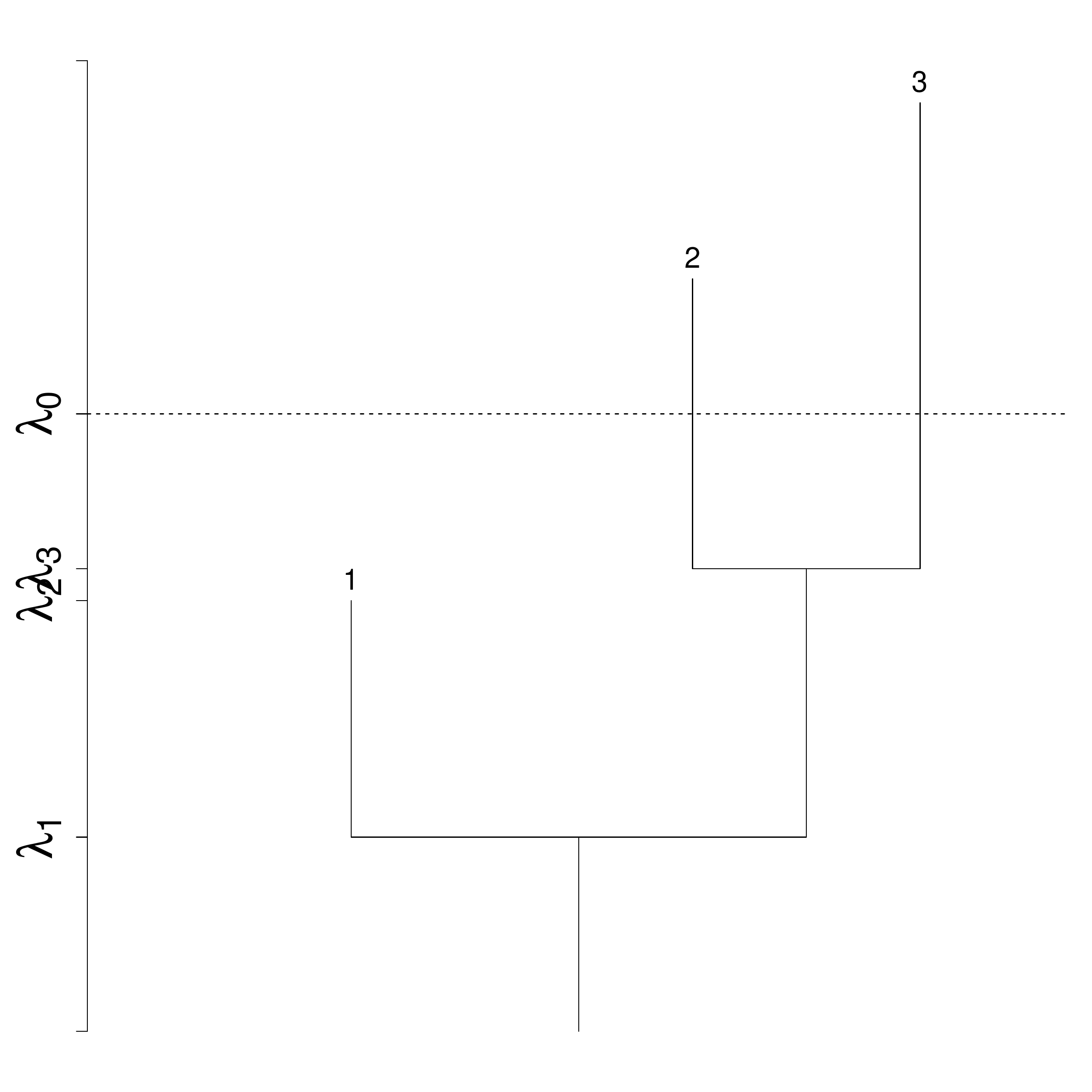}
\end{tabular}
\end{center}
\vspace{-.4cm}\caption{A section of a density function at a level $\lambda_0$ (left), the identified level
set (middle panel), formed by two disconnected regions and the associated
cluster tree, with leaves corresponding to the modes. The horizontal line is at the level $\lambda_0$ (right).} \label{fig:modal}
\end{figure*}

Nonparametric, or modal, clustering hinges on the assumption that the observed data $(x_{1}, \ldots, x_{n})'$ are 
a sample from a probability density function $f:\mathbb{R}^{d} \mapsto \mathbb{R}^{+}$.
The modes of $f$ are regarded as the archetypes of the clusters,
which are in turn represented by the surrounding regions.
The identification of the modal regions is performed according to two alternative directions.  
One strand of methods looks for an explicit representation of the modes of
the density and associates each cluster to the set of points along the steepest ascent path towards a mode.
Optimization method are applied to find the local maxima of the density, such as
the \emph{mean-shift} algorithm \cite{Fukunaga_Hostetler:75},
and a number of its variants \cite{meanshift1, meanshift2, meanshift3}.
We consider, alternatively, a second strand, which associates the clusters to
disconnected density level sets of the sample space, without attempting the explicit task of mode
detection. Specifically, a section of $f$ at a given level $\lambda$ singles
out the (upper) level set
$$L(\lambda) = \{x\in \mathbb{R}^d: f(x)\geq \lambda\}, \quad  0 \leq \lambda \leq  \max f$$
which may be connected or disconnected. In the latter case, it consists of a number of connected components, each of them 
associated with a cluster at the level $\lambda$.

While there may not exist a single $\lambda$ which catches all the modal regions,
any connected component of $L(\lambda)$ includes at least one mode of the density and, on the other hand, for each mode 
there exists some $\lambda$ for which one of the connected components of the associated $L(\lambda)$
includes this mode at most. Hence, not only it is not necessary to define a specific level $\lambda$ to identify the groups, which would be difficult and 
often not effective in providing the overall number of modes, but conversely,
all the modal regions may be detected
by identifying the connected components of $L(\lambda)$ for different $\lambda$s. 
Varying $\lambda$ along its range gives rise to a hierarchical structure of the high-density sets,
known as the \emph{cluster tree}. For each $\lambda$, it provides the number of connected components of $L(\lambda),$ 
and each of its leaves corresponds to a \emph{cluster core}, \emph{i.e.}
the largest connected component of $L(\lambda)$ including one mode only. 
Figure \ref{fig:modal} illustrates a simple example of this idea: cluster cores associated
with the highest modes $2$ and $3$ are identified by the smallest $\lambda$ larger than $\lambda_3$,
while the smallest $\lambda$ larger than $\lambda_1$ identifies the cluster core associated to mode $1$. In some sense, 
the cluster tree provides a tool to inspect data with different degree of sharpness: clusters 2 and 3 are distinct, but they merge together to create a lower resolution cluster. 
Instead of indexing the tree with $\lambda,$ it is equivalent to consider the probability associated to $L(\lambda)$, which varies inversely with $\lambda$. For this reason the tree
is usually plotted upside down. This will be also the convention considered for the rest of the paper.

From the operational point of view, two choices are required to implement the ideas underlying nonparametric clustering.  
Since $f$ is unknown, a nonparametric estimator $\hat{f}$ is employed to obtain its representation. A common
choice for multidimensional data is the product kernel estimator \cite[see, e.g., ][]{Scott_Sain05}:
\begin{equation}\label{f}
\hat{f}(x; h) = \sum_{i=1}^n \frac{1}{n h_1\cdot \ldots \cdot h_d}  \prod_{j=1}^d K \left(\frac{x^{(j)} - x_i^{(j)}}{h_j} \right)
\end{equation}
where $x^{(j)}$ is the $j-th$ component of $x$, the univariate kernel $K$ is usually taken to be a unimodal, nonnegative function centered at
zero and integrating to one, and a different smoothing parameter $h_j$ is chosen for each component.
In fact, for the development of the method, it does not really matter which specific estimator is adopted,
provided that $\hat{f}$ is positive and finite at the observed points.

A second choice derives from the lack, in multidimensional sample spaces, of an obvious method to identify
the connected components of a level set. For these reasons the inherent literature has mainly focused on developing efficient methods
for this task  \citep[\emph{e.g.} ][]{Stuetzle_Nugent10, Menardi_Azzalini13}. 

Note that the union of the cluster cores does not produce a partition of the sample space, as regions at the tails or at the valleys of $f$, where the attraction of each mode is low, are initially left
unallocated. However, the availability of a density measure allows for providing each unallocated observation with a degree
of confidence of belonging to the cluster cores. Depending on the application, the evaluation of such confidence
may be exploited to force the assignment or may result in the opportunity of fuzzy clustering schemes.

\subsection{Related works on image segmentation}
Also due to the extensiveness of its applicability, several different methods have been proposed to
pursue the task of image segmentation. These are broadly ascribable to two alternative routes \cite[see, for a review, ][]{Dougharty09}: \emph{noncontextual}
techniques ignore the relationships among the features in the image and assign the same label to pixels sharing some global attribute,
such as the gray level or the color brightness. \emph{Thresholding}, for instance, compares the intensity of each pixel with a
suitable threshold and associates higher values to the foreground of the image, of main interest, and lower values to the background. Recent contributions within this framework are \cite{Ayala_etal15, elAziz_etal17}. 
\emph{Contextual} techniques, conversely, also account for pixel location or color gradient. 
Within this class, \emph{region-based methods} mainly rely on the assumption that the
neighboring pixels within one region have similar value \cite[see, e.g.][]{chan_vese01, li_etal08, ning_etal2010}. Boundary-based methods as \emph{edge detection} and \emph{active contours}
build on finding pixel differences rather than similarities,
to determine a closed boundary between the foreground and the background of the image. Examples of recent procedures following these approaches are \cite{kimmel03, dollar13} and, respectively, \cite{xiang_etal06, zhang_etal17}.  
\emph{Watershed} segmentation builds a distance map of a gray-scale image or of its gradient and considers it as topographic relief,
to be flooded from its minima. When two lakes merge, a dam is built, representing the boundary between two segments \cite[e.g.][]{seal_etal15}.

Within the framework of clustering methods, $K-$means clustering is diffusely used for image segmentation \cite{sulaiman10},
perhaps due to its simplicity, but a few severe limitations prevent its effectiveness.
First, $K-$means clustering is known to produce sub-optimal solutions as it highly depends
on the initialization of the centroids. Additionally, it requires a prior specification
of the number of clusters. In image segmentation this operation is undoubtedly easier than
in other clustering applications. On the other hand, the need of human intervention vanishes
the effort to automate the segmentation procedure. Finally, $K-$means is known to be biased
toward the identification of spherical clusters, which can be restrictive in
image data where segments may assume arbitrarily odd shapes. 

While nonparametric clustering is rarely mentioned as the underlying approach to perform image analysis, it 
features some connection with a number of segmentation algorithms. By exploiting some notions from Morse theory, in
\cite{Chacon15} an elegant formalization of the 
notion of modal (nonparametric) cluster, which closely recalls the ideas of watershed segmentation is provided:
intuitively, if the density underlying the data is figured as a mountainous landscape, and modes are its peaks, clusters
are the ``regions that would be flooded by a fountain emanating from a peak of the mountain
range''. Furthermore, {segmentation \emph{thresholding} is often performed by looking for the minimum of the histogram of the grey intensities, i.e. the antimode of a histogram-based density estimator of the grey intensities. Since any antimode lies between two modes, the approach
can be interpreted as a naive, single-$\lambda$, implementation of the density level set formulation above mentioned,
where gray intensities are employed as a measure of density.}
While without a specific reference to nonparametric clustering, gradient ascent algorithms in the guise of mean-shift, have been also sometimes applied  
for image segmentation \citep{Tao_etal07, liu08, zhou_etal13}. Indeed, by climbing the modes of a kernel density estimate, the mean-shift is rightly ascribable to the class of non parametric clustering methods. Similar instruments are also at the basis of active contours models,
where a suitable measure of energy is iteratively minimized by a gradient descent algorithm to identify the
segment contours.

As a further link, even when applied to different goals, image analysis and nonparametric clustering share several tools:
an example is provided by spatial tessellation as the Voronoi or Delaunay diagrams, which have been used in nonparametric clustering to
identify density level sets connected components \cite{Azzalini_Torelli07} and are frequently employed in image analysis for thinning and skeletonization.

\section{A nonparametric method for image segmentation}
\subsection{Description of the procedure}\label{sec:procedure}
Let $\mathcal{I} = \{p_1, \ldots, p_n\}$ be a digital image, where the ordered set of pixels
$p_i = ((x_i, y_i), z_i), i= 1, \ldots, n,$ is described by the pair $(x_i, y_i)$ denoting the coordinates of the pixels location, and by the vector $z_i$
denoting the adopted color model, \emph{e.g.} $z_i = (z_i^{(r)}, z_i^{(g)}, z_i^{(b)})$ in the RGB color representation \cite{Soille13}.
In grayscale images, $z_i$ is a scalar quantifying the gray intensity. 

The particularization of nonparametric clustering in the framework of image analysis requires a density function to be
defined at the pixels. A sensible choice builds $\hat{f}$ based on the color coordinates $z_i$.
The specification of the \eqref{f} is then:  

\begin{equation}\label{eq:fcol}
\resizebox{0.9\hsize}{!}{$
\hat{f}(z) = \frac{1}{nh_rh_gh_b} \sum_{i=1}^n K \left(\frac{z^{(r)} - z_i^{(r)}}{h_r}\right) K \left(\frac{z^{(g)} - z_i^{(g)}}{h_g}\right) K \left(\frac{z^{(b)} - z_i^{(b)}}{h_b}\right).
$}
\end{equation}

For example, if the Uniform kernel $K(u)=\tfrac{1}{2}\mathbf{1}_{\{|u| < 1\}}$ is selected and $h_j \to 0$, each pixel is provided
with a density proportional to the frequency of its color in the image. 
Similar interpretations hold with different kernel functions. 

\begin{figure}[tb]
\begin{center}
\hspace{.4cm}\includegraphics[width=.3\textwidth]{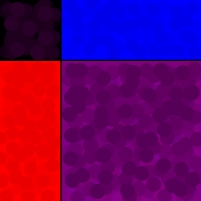}\\
\vspace{.5cm}\includegraphics[width=.32\textwidth]{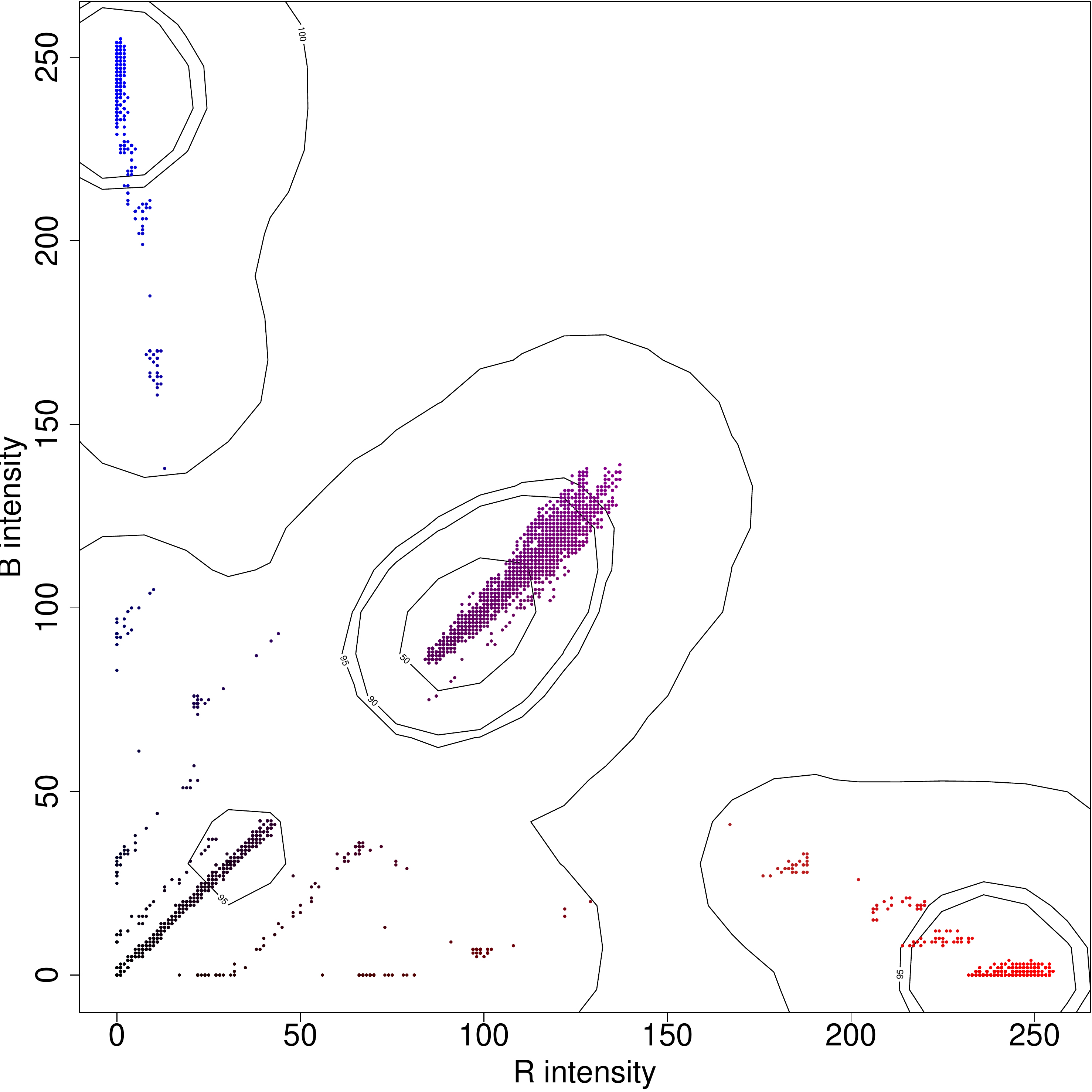}
\end{center}
\caption{A 4-segments toy image entirely characterized by red and blue channels (top); red and blue intensities of the pixels and, superimposed, their kernel density estimate, highlighting 4 clear modes corresponding to the 4 segments (bottom).}\label{fig:RB} 
\end{figure}

Consider, as an example, the image in the top panel of Figure \ref{fig:RB}. For the sake of illustration, the image has been built by using colors entirely characterized by the red and blue RGB channels (\emph{i.e.} each pixel has the green channel set to 0). The image is clearly composed by 4 segments, each of them featured by a main color pattern appearing with different shades. The bottom panel displays the color intensities of each pixel, disregarding their location within the image. Pixels sharing a similar color have similar red and blue intensities, and cluster together in different areas of the plane. Hence, the color density estimate exhibts 4 modes, associated to the 4 color patterns.

As it will be discussed in Section \ref{sec:discussion}, an alternative to \eqref{eq:fcol} would also account for
the spatial coordinates of the pixels. 

Once that the color density has been estimated, the associated upper level sets
$$\hat{L}(\lambda) = \{p_i\in \mathcal{I}: \hat{f}(z_i)\geq \lambda\}\qquad 0 \leq \lambda < \max \hat{f}$$
are easily determined for a grid of $\lambda$ values. Next step is the identification of the connected components of the $\hat{L}(\lambda)'$s. 
Unlike the above mentioned case of clustering data on $\mathbb{R}^d$, where the identification of connected regions
is ambiguous, the notion of connectedness is (almost) unequivocally determined in the image framework, due to the spatial structure
of the pixels. 
This justifies the procedure here proposed, which builds on the level set formulation of nonparametric
clustering but naturally exploits the adjacency structure of the pixels to identify the connected
components of the modal regions.
For a given $\lambda$, the connected components of $\hat{L}(\lambda)$ are approximated as follows:
\begin{itemize}
\item[(i)] for each pixel of $\hat{L}(\lambda)$, identify the adjacent pixels forming its \emph{4-neighbourhood}, 
\emph{i.e.} a central pixel has four connected neighbors - top, bottom, right and left.
\item[(ii)] approximate the connected components of $\hat{L}(\lambda)$ by the union
of adjacent pixels in $\hat{L}(\lambda)$.
\end{itemize}

Hence, in order to identify a segment, not only the pixels need to share a similar color, but a constraint of spatial connectedness is also imposed by the procedure. For example, 
if a black spot was embedded in the violet segment in the toy image of Figure \ref{fig:RB}, the associated pixels would contribute to form (and hence raise) the mode of the density at the bottom corner of the lowermost panel of the Figure. However, at a level $\lambda$ for which the black colored pixels would have higher density, $\hat L(\lambda)$ would be disconnected, because formed by both the pixels of the black square already in the image, and the pixels of the embedded, not adjacent, black spot.

\begin{algorithm}[tb]
\caption{Nonparametric density-based segmentation: main steps of the procedure}\label{pseudocode}
\begin{algorithmic}[1]
\Require $h = (h_r, h_g, h_b)$; $K(\cdot)$; $\epsilon$; 
 Class $\in$ \{TRUE, FALSE\} (set Class:= FALSE to not allocate pixels not belonging to the segment cores;)
\State Identify the set $N_4(p_i)$ of pixels in the 4-neighborhood of $p_i, i = 1, \ldots, n$
\State Compute $\hat{f}(z) = \sum_{i=1}^n \frac{1}{n h_r h_g h_b} \prod_{j=\{r,g,b\}} K \left(\frac{z^{(j)} - z_i^{(j)}}{h_j} \right)$,  $\quad \forall z \in \{z_i\}_{i=1, \ldots, n}$
\State Set $\lambda: = 0$
\While{$0 \leq \lambda \leq \max\hat{f}$}
\State identify $\hat{L}(\lambda)=\{p_i: \hat{f}(z_i)\geq \lambda\}$;
\State find the connected components of $\hat{L}(\lambda)$ (as the union of adjacent pixels in $\hat{L}(\lambda)$)
\State $\lambda: = \lambda + \epsilon$
\EndWhile
\State Build the hierarchy of the connected components of $\hat{L}{(\lambda)}'s$ and obtain the cluster tree  
\State Denote core pixels as $p_c$ and unallocated pixels as $p_{u}$
\State Assign the label $\ell_c \in \{1, \ldots, M\}$ to each core pixel $p_c$ 
\State Set Isolated:= $\emptyset$
\If {Class = TRUE}
\While{$\{p_u\} \neq \emptyset \,\, \vee $ \{Isolated\} $= \{p_u\}$ }
\State Set Isolated:= $\emptyset$
\For {all $p_u$} 
\State $\hat{f}_m (z_u) = \sum_{c: \ell_c=m} \frac{1}{n h_r h_g h_b} \prod_{j=\{r,g,b\}} K \left(\frac{z^{(j)} - z_c^{(j)}}{h_j} \right)$,
  $m=1 \ldots,M$ 
\State set $m_0 := \mbox{argmax}_m \hat{f}_m(z_u)$ 
\If {$\exists$ $p_c$ such that ($p_c$ $\in$ $N_4(p_u) \wedge \ell_c = m_0$)}
\State assign the label $\ell_c:=m_0$ to $p_u$
\Else 
\State Isolated: = \{Isolated $\cup \,p_u$\} 
\EndIf
\EndFor
\State update $\{p_u\}$ and $\{p_c\}$
\EndWhile
\Else 
\State set $\ell_u:= 0$

\EndIf\State {\bf RETURN: $\ell_1, \ldots, \ell_n$} 
\end{algorithmic}
\end{algorithm} 

For varying $\lambda$, the procedure described so far creates $M$ groups of pixels $\ell_m$ ($m = 1, \ldots, M$), which we call, in analogy with the clustering problem,
(segment) \emph{cores},  
and it leaves a number of pixels 
unlabeled. Depending on the application at hand, we can either decide to force their assignment to the existing
segment cores or to leave these pixels unallocated. In fact, a peculiar aspect is that the unlabeled 
points are not positioned randomly in the image, but are inevitably on the outskirts of the existing
segment cores. As will be illustrated in the Section \ref{sec:examples}, unallocated pixels include (or correspond to) the
contours of the segments.   

The possible allocation of the unlabeled pixels to the existing groups is essentially a classification problem that may be faced
according to a wide range of choices. To remain within the same kind of approach pursued so far, and consistently with the
purpose of identifying segments as connected sets of pixels, we propose the following idea to classify an unallocated pixel $p_u$:
 compute the $M$ estimated densities $\hat{f}_m(z_u)$, each based on the pixels 
already assigned to the $m^{th}$ core only ($m = 1,\ldots, M)$; then, 
set 
\begin{equation}\label{eq:unallocated}
m_0 = \mbox{argmax}_m \hat{f}_m(z_u) 
\end{equation} 
and assign $p_u$ to the group with label $m_0$ provided that at least one of the pixels already assigned to the $m_0^{th}$ segment core
is adjacent to $p_u$. The operational implementation of this idea is here performed in a sequential
manner, as detailed in the reported pseudo-code, along with the main steps of the whole segmentation procedure.

It may be the case that none of the pixels already assigned to the segment presenting maximum density (\ref{eq:unallocated}) is adjacent to $p_u$, (\emph{i.e.} the color of $p_u$ is not similar to the color of any other adjacent segment). These pixels usually lie at the borders of the segments and may either be left unallocated, or the assignment can be forces to the highest density segment, disregarding spatial adjacency, or to a novel further segment. 

\begin{figure*}[tb]
\begin{center}
\includegraphics[height=1.6in, width=.73\textwidth]{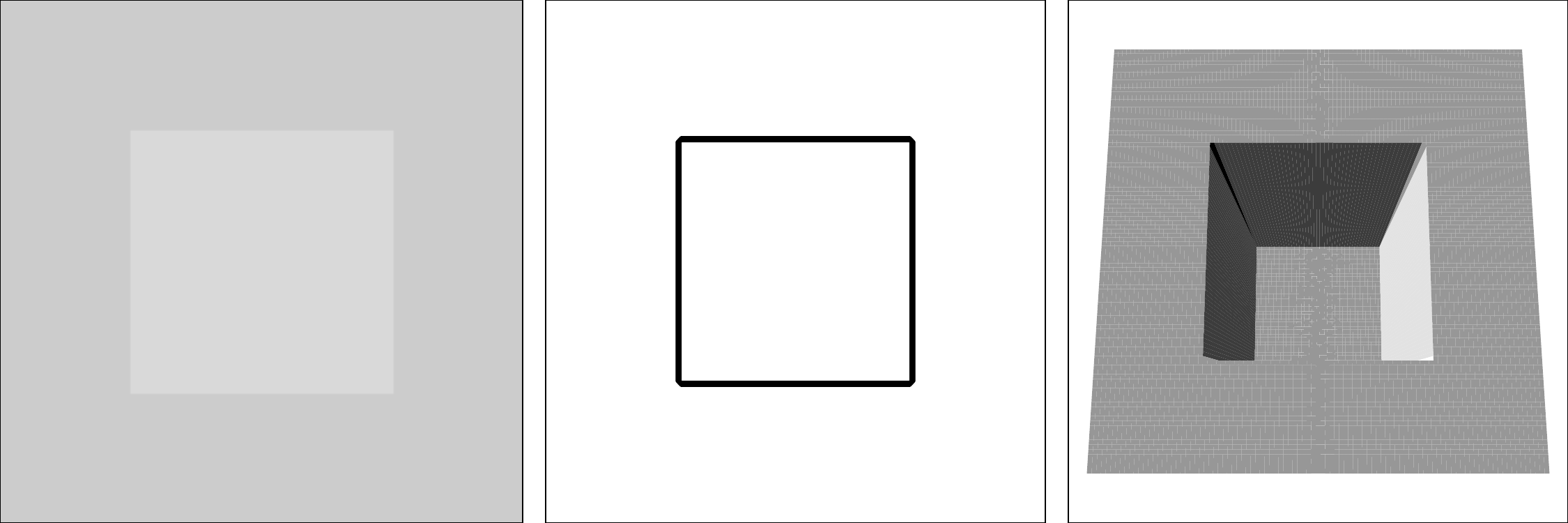}
\end{center}
\caption{A simple gray-scale image (left), and the estimate of color density superimposed to the spatial coordinates to the pixels,
represented both as the level set density (middle panel) and as a perspective plot (right panel).}\label{fig:quadrato} 
\end{figure*}

\subsection{Discussion}\label{sec:discussion}

\begin{figure*}[bt]
\begin{center}
\fbox{\includegraphics[height=1.6in]{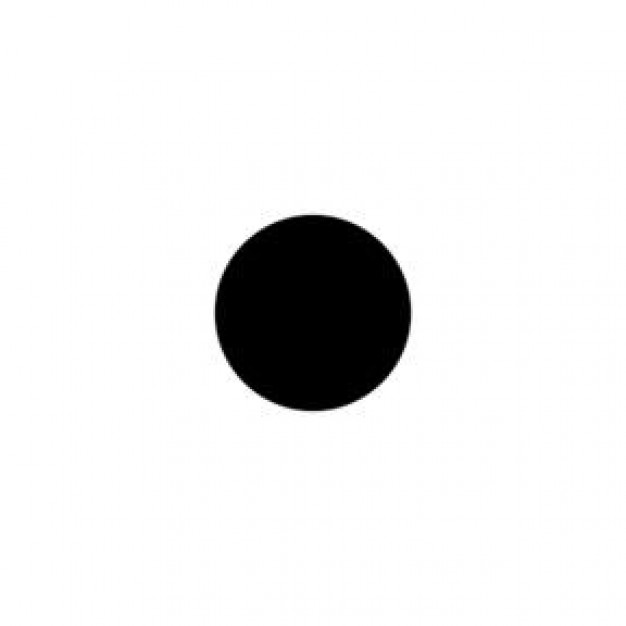}}
\hspace{.1cm}
\fbox{\includegraphics[height=1.6in]{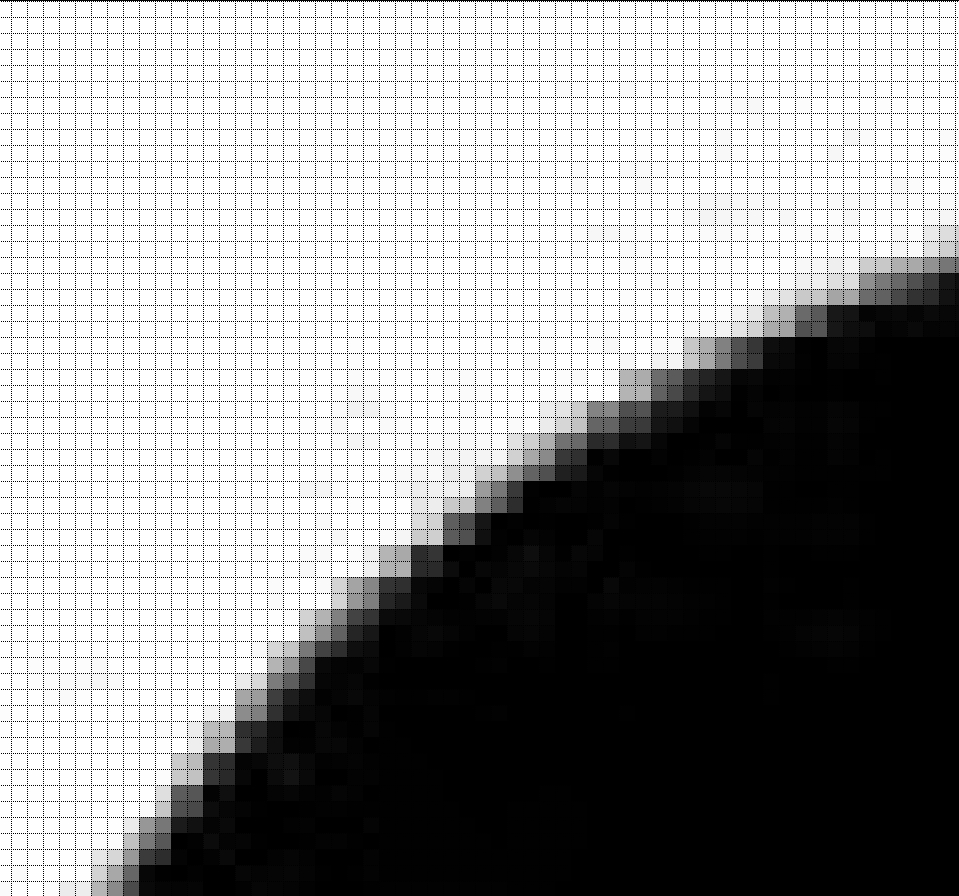}}
\includegraphics[height=1.7in]{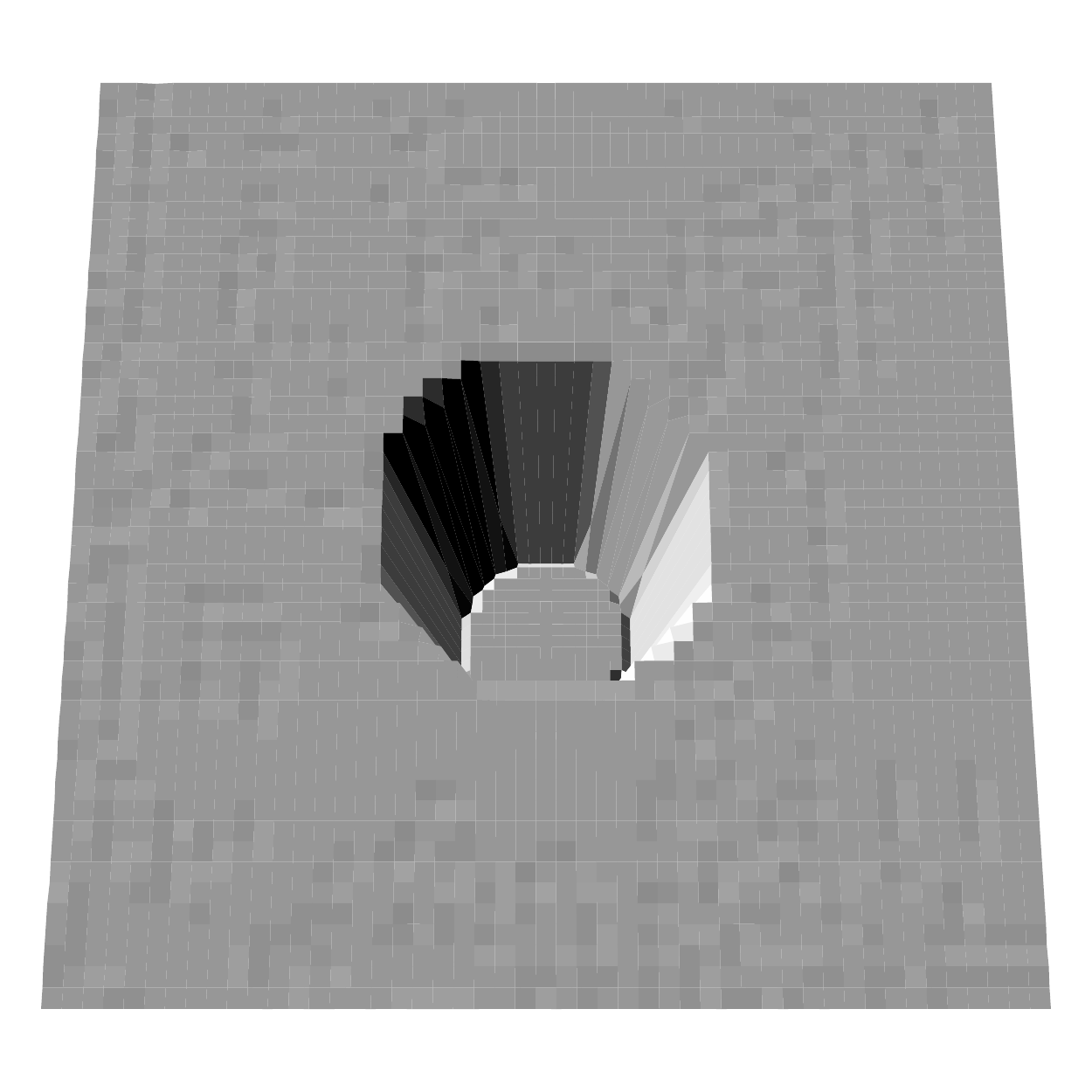}
\end{center}
\caption{A simple gray-scale image (left), and a zoommed detail (middle), showing that
when a segment has either curve or sloped contours, the color is ripped along the border,
to prevent a sawtoothed rendering. Due to this feature, the perspective plot of the density estimate, based on
the \eqref{eq:fcol} highlights a valley at the border of the foreground (right).}\label{fig:cerchio1} 
\end{figure*}

Since the procedure illustrated so far accounts for both the colors and the connectivity of the image patterns,
it emulates, in some sense, the behavior of the human eye, which instinctively, perceives different segments in a picture as
either disconnected set of pixels or image patterns with a diverse color intensity.
A simple illustration of this latter aspect is witnessed by the grayscale image in Figure \ref{fig:quadrato}.
Even if the gray intensities of the foreground (the inner square) and the background are similar, the density estimator \eqref{eq:fcol} 
perfectly distinguishes the two density levels and the isoline identifies the contours of the foreground segment. 
Conversely, with respect to the former aspect, a major limitation of nonparametric clustering,
in principle inherited by the proposed segmentation procedure, derives from the definition of mode itself, which requires
 a separating ``gap'' between dense patterns. 
In Figure \ref{fig:quadrato}, the density shows itself like a squared hole, and there is no lower
density area between the background and the foreground. This prevents $\hat{L}(\lambda)$ to be a disconnected set for any $\lambda$, 
which would guarantee the identification of two segments. This behavior is somewhat paradoxical, as the neater the image,
the less ideal the setting for the procedure to work effectively: within an image, dripped contours of a segment, indeed, manifest themselves
as small changes of colors at the borders with respect to the interior. Since the perimeter of a shape is always smaller
than its area, and the density of a pixel is positively associated with the frequency of its color, dripped contours would
guarantee that the color density along the contour of a segment is lower than its inner density,
and hence a valley would arise between a segment and its background.
In fact, the considered example has been built ad hoc by setting the gray intensity for each pixel. In practice, many images
have segment contours not defined with such neatness, no matter what the image resolution is.
This is especially true with segments having either curve or sloped contours, since the shades of colors along the border 
of the segments allow to prevent a sawtoothed rendering. See Figure \ref{fig:cerchio1} for an illustration.

When the image does not features itself with dripped contours, it is possible to overcome the issue of lacking valleys in
 the density by introducing some blurring of the image. To this aim, given that the identification of pixel neighbors
 is required anyway for the identification of disconnected regions, a simple strategy is to replace each pixel value
 with an average of the values in its neighborhood. In fact, a quite common practice prior to thresholding segmentation is
 to smooth an image using averaging or median mask. See \cite[][\S 10.2.1]{Dougharty09}. 
 
A further, somewhat related, issue concerns the choice of the density measure. While we choose to build $\hat{f}$ based on the color
 intensities only, an alternative route would consist in exploiting the whole available information in terms of both the color coordinates
 and the spatial coordinates, \emph{i.e.}: 
\begin{equation}\label{eq:fall}
\resizebox{0.91\hsize}{!}{$
\hat{f}(p) = \frac{1}{n\prod_j h_j} \sum_{i=1}^n K \left(\frac{z^{(r)} - z_i^{(r)}}{h_r}\right) K \left(\frac{z^{(g)} - z_i^{(g)}}{h_g}\right) K \left(\frac{z^{(b)} - z_i^{(b)}}{h_b}\right) K \left(\frac{x - x_i}{h_x}\right) K \left(\frac{y - y_i}{h_y}\right).
$}
\end{equation}
Note that, depending on the choice of  both the kernel and the bandwidth, the \eqref{eq:fall} can be a function of the distance between $(x, y)$ and $(x_i , y_i)$. 
Let, for instance, $h_x = h_y = 1.$ Then, the last two factors are proportional to $e^{d_2((x,y), (x_i, y_i))^2}$ or to $e^{d_1((x,y), (x_i, y_i))}$ when a Gaussian or, respectively, a Laplace kernel is selected, with $d_p(\cdot, \cdot)$ the $L_p$ distance.  

\begin{figure}[bt]
\begin{center}
\includegraphics[height=5.6cm, angle =-90]{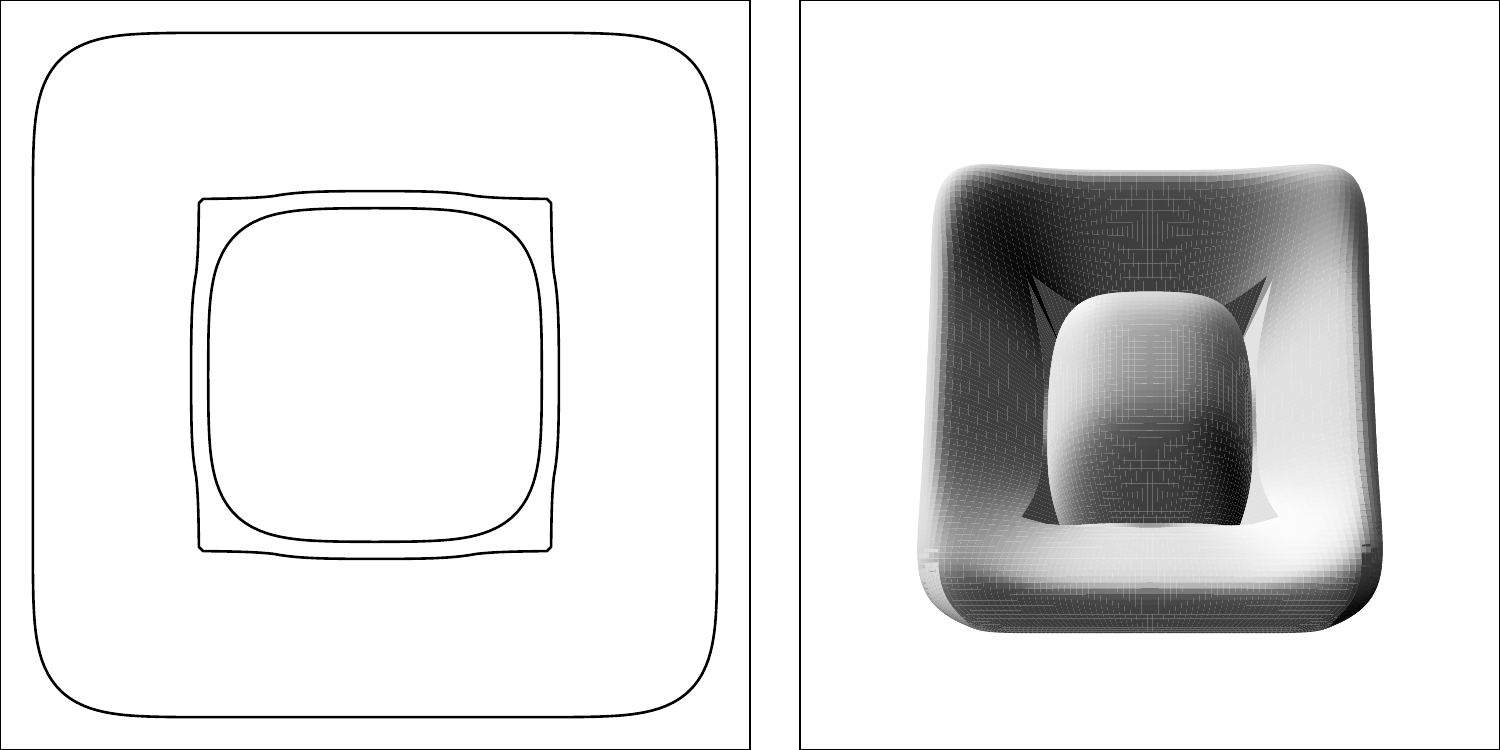}
\end{center}
\caption{Density estimate superimposed to the grayscale image in Figure \ref{fig:quadrato} when $\hat{f}$ is based both on the
colors and the spatial coordinates: level set representation (top panel) and
perspective plot (bottom panel).}\label{fig:quadratocoo} 
\end{figure}

Estimating the density as in \eqref{eq:fall} would also overcome the above mentioned problem of lacking valleys at the borders of
the segments: in \eqref{eq:fcol}, the largest contribution to the density of a generic pixel is provided by     
all the pixels having similar colors; conversely, if also the spatial coordinates are involved in $\hat{f}$, the
density of a generic pixel depends on pixels with similar colors \emph{and} close spatially. Hence,
at the borders of a segment, where part of the adjacent pixels have a different color,
the density turns out to be lower than the interior pixels (see Figure \ref{fig:quadratocoo} for an illustration).
While this behavior is desirable for the purpose of segmentation, the interpretation of $\hat{f}$ in terms of color
frequency fails and, indeed, a higher computational effort is required. Additionally, some empirical work not included in the manuscript
has proven that estimating $f$ via the \eqref{eq:fall} results in oversegmenting the image, hence using the only color coordinates to estimate density 
is overall preferable.

In fact, two further aspects concerning the estimation of $\hat{f}$ need to be accounted for, concerning the selection of the kernel function $K$ and the smoothing vector $h$. With respect to the former choice, it has been well-established that it does not have a strong impact on the density estimate \cite[see, e.g.][]{Wand_Jones94b, Chacon_Duong18}. However, the use of a bounded-support kernel, such as the uniform one (which would be overall more easily interpretable in this context) is in general not advisable for image segmentation, as it entails that in the evaluation of the density of a pixel, other pixels have weight either constant and positive, or null, depending on $h$ and on the difference between color intensities. 
In this sense, different hues of the same colors could be evaluated as either the same colors or as completely different colors. For this reason, and especially when colors are not homogeneous within the image, a longer-tail kernel is more secure. Concerning the choice of $h,$ on the other hand, in clustering it is not as critical as it is in density estimation, since a rough indication about the high density location may suffice. To this aim, one may resort to one of the numerous bandwidth selectors proposed in literature about kernel density estimation, such as the ones based on cross-validation or plug-in strategies. The reader may refer to \cite{Chacon_Duong18} for a recent review.  
In any case, both choices are certainly issues to be tackled, and will represent object of empirical investigation in the next section.    
   
As a final note of this section, observe that, since the proposed procedure relies on the definition of a segment as a cluster of connected pixels sharing the same color intensity, it cannot identify segments which configure themselves as isolated pixels within some differently colored background. Depending on the application, and on the image resolution, this may represent a limitation, or not. For example, the stars in the image of a starry sky might appear as single isolated yellow pixels if the image has low resolution (\emph{i.e.} it is small), thus likely required to be identified as small segments. On the other hand, single differently-colored pixels might be just spots or imperfections, which would be preferably not segmented.   

\section{Empirical work}\label{sec:examples}

\subsection{Simulation study}
{As a first step of the empirical analysis, we present the results of a compendious simulation study aimed at understanding and systematizing the strengths and the limits of the proposed segmentation method, also with respect to the sensitivity of the density function to different kernels and smoothing amounts.

To provide the images with the randomness required to run simulations, yet guaranteeing a given segmentation structure, images are generated as follows: each pixel is associated a priori with a nominal color intensity, which defines the segment membership; then, the actual intensity is randomly drawn from a normal distribution having the nominal intensity as a mean. 

Four main simulation settings are considered. Stemming from a reference image, with two convex shaped segments and no clefts (setting A1), one characteristic is changed at a time for each simulation setting, in order to control for its effect on the resulting segmentation. A larger number of smaller segments is considered (setting B1), non-convex shapes (setting C1), as well as the presence of clefts in the image, where the color intensity might mix up with the color intensity of neighbour segments (setting D1).     
Within each of these settings, different degrees of color homogeneity are simulated by increasing the variance of the normal distribution from which the color intensity of each pixel is drawn (setting A2, B2, C2, D2).  Additionally, shaded segment contours are considered, determined by setting the border color intensity at the mean of the nominal intensities of the adjacent segments (setting A3, B3, C3, D3). One example of image generated from each setting and subsetting is displayed in the top of Tables} \ref{settingA}, \ref{settingB}, \ref{settingC}, \ref{settingD}.

\begin{table}[b!]
\caption{Simulation results for setting A: each entry displays the Monte Carlo average of the Adjusted Rand Index (ARI) and of the number of detected segments (the true number of segments is 2). Standard deviations are reported in brackets. }\label{settingA}
\centering
\resizebox{.49\textwidth}{!}{%
\begin{tabular}{p{1.8cm}p{1.55cm}p{2cm}p{1.9cm}p{2cm}}
\hline
&  &  \includegraphics[scale=2]{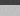} \newline A1: benchmark  & \includegraphics[scale=2]{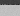} \newline A2: shaded contours & \includegraphics[scale=2]{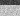} \newline A3: heterogeneous colors \\   \hline\hline
         &ARI & 0.99 & 0.98 & 0.91 \\ 
Normal K &    & (0.08) & (0.03) & (0.14) \\ 
$h=h_N$&  \# segments & 3.10 & 2.92 & 2.71 \\ 
&   & (1.54) & (1.52) & (1.22) \\ \hline
&  ARI & 1.00 & 0.99 & 0.60 \\ 
Normal K&  & (0.00) & (0.02) & (0.20) \\ 
$h=0.75h_N$&  \# segments & 2.00 & 2.00 & 5.60 \\ 
&   & (0.00) & (0.09) & (2.22) \\ \hline
&  ARI & 1.00 & 0.96 & 0.77 \\ 
Normal K&  & (0.09) & (0.06) & (0.22) \\ 
$h=1.25h_N$&  \# segments & 2.00 & 2.27 & 4.07 \\ 
&   & (0.00) & (0.83) & (2.22) \\ \hline\hline
&  ARI & 1.00 & 0.54 & 0.57 \\ 
Uniform K&   & (0.00) & (0.45) & (0.21) \\ 
$h=h_N$&  \# segments & 2.00 & 1.86 & 5.68 \\ 
&   & (0.00) & (1.08) & (2.06) \\ \hline
&  ARI & 1.00 & 0.95 & 0.66 \\ 
Uniform K&  & (0.00) & (0.11) & (0.23) \\ 
$h=0.75h_N$&  \# segments & 2.00 & 2.18 & 4.92 \\ 
&   & (0.00) & (0.73) & (2.33) \\\hline
&  ARI & 0.55 & 0.48 & 0.48 \\ 
Uniform K&   & (0.16) &(0.38) & (0.16) \\ 
$h=1.25h_N$&  \# segments & 6.72 & 2.82 & 6.55 \\ 
&   & (1.86) & (1.84) & (1.78) \\\hline\hline
\end{tabular}}
\end{table}

\begin{table}[b!]
\caption{Simulation results for setting B: each entry displays the Monte Carlo average of the Adjusted Rand Index (ARI) and of the number of detected segments (the true number of segments is 9). Standard deviations are reported in brackets. } \label{settingB}
\centering
\resizebox{.49\textwidth}{!}{%
\begin{tabular}{p{1.8cm}p{1.55cm}p{2cm}p{2cm}p{2cm}}
\hline
&  &  \includegraphics[scale=2]{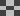} \newline B1: benchmark   & \includegraphics[scale=2]{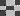} \newline B2: shaded contours  & \includegraphics[scale=2]{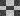} \newline B3: heterogeneous colors   \\ 
 \hline\hline
&ARI & 0.99 & 0.83 & 0.87 \\ 
Normal K& & (0.02) & (0.04) & (0.10) \\ 
$h=h_N$ &\# segments & 9.51 & 9.31 & 8.75 \\ 
  & & (0.70) & (0.79) & (1.36) \\\hline 
&ARI & 0.99 & 0.84 & 0.93 \\ 
 Normal K& & (0.02) & (0.02) & (0.06) \\ 
 $h=0.75h_N$ &\# segments & 9.16 & 9.19 & 9.95 \\ 
  & & (0.39) & (0.44) & (1.18) \\ \hline
  &ARI & 0.84 & 0.78 & 0.82 \\ 
  Normal K& & (0.13) & (0.08) & (0.10) \\ 
  $h=1.25h_N$ &\# segments & 6.99 & 9.00 & 8.66 \\ 
  & & (1.13) & (1.34) & (1.53) \\ \hline\hline
  &ARI & 1.00 & 0.78 & 0.91 \\ 
  Uniform K& & (0.02) & (0.06) & (0.07) \\ 
  $h=h_N$ & \# segments &8.98 & 9.30 & 9.68 \\ 
  & & (0.20) & (0.98) & (1.23) \\\hline
  &ARI & 0.65 & 0.78 & 0.96 \\ 
  Uniform K& & (0.00) & (0.07) & (0.04) \\ 
  $h=0.75h_N$ & \# segments &5.00 & 8.97 & 9.34 \\ 
  & & (0.00) & (0.83) & (0.64) \\ \hline
  &ARI & 0.94 & 0.65 & 0.85 \\ 
  Uniform K& & (0.11) & (0.12) & (0.09) \\ 
  $h=1.25h_N$ &\# segments & 8.48 & 7.99 & 8.96 \\ 
  & & (1.42) & (1.54) & (1.50) \\ \hline\hline
\end{tabular}}
\end{table}

{For the analyses, density is estimated via }\eqref{eq:fcol}, {with both Gaussian and Uniform kernels. 
The bandwidth $h$ is selected as the asymptotically optimal solution for data following a Normal distribution (in the following, $h=h_N$). This selection criterion is, by construction, sub-optimal. Indeed, any assumption of Normality of the data color distribution cannot hold since multi-segment images have, in principle, a multimodal density. Nevertheless, this rule of thumb has resulted quite effective in clustering applications and has been then used in this analysis due to its simplicity and low computational burden. In order to understand the sensitivity of the procedure to different amounts of smoothing,  the bandwidth has been both shrinked and enlarged slightly to the values $h = 0.75h_N$ and $h = 1.25h_N$ to evaluate the effect on the segmentation. 

The agreement between the true and the detected segment membership has been measured in terms of number of detected segments and Adjusted Rand Index }\cite{ari}, {which takes increasing values for improved quality of the segmentation (the maximum value $1$ is associated to a perfect segmentation).     

In each simulation setting 500 grayscale images of $320$ pixels are generated, as this allows for keeping feasible the computational burden.} All the empirical work has been performed in the \verb,R, environment \cite{R}. Images have been handled via the packages \verb,EBIimage, \cite{EBImage} and \verb,BiOps, \cite{biops}, while the segmentation routines have been built as adjustments of 
the clustering routines available in the package \verb,pdfCluster, \cite{pdfCluster}.  

{Simulation results are summarized in Tables} \ref{settingA}, \ref{settingB}, \ref{settingC}, \ref{settingD}. {The procedure works comprehensively in a satisfactory way. With a Normal kernel, it provides almost perfect segmentations in the simplest setting A1, A2, A3, where results are robust to different amounts of smoothing. In the more challenging settings featured by shaded contours and heterogeneous color segments the performance of the segmentation worsens, yet remaining remarkably good in most of the considered frameworks.      
Neither non-convex segment shapes or the presence of many small segments do compromise the segmentation. In the latter case, segmentation may worsen for large smoothing, due to the risk of cluster merging when the density is oversmoothed. The greatest challenge is represented by the setting D (Table }\ref{settingD}), {since within interstices of the segments (like the thin frame of the main image) the color intensity mixes up with the intensity of the neighbours segments and prevents the edge detection. Increasing the smoothing amount may help to reduce the entailed oversegmentation. In fact, this behaviour is exacerbated by the small size of the segmented images, as thin clefts in high-resolution images can be anyway associated to relatively large spatial areas, and the color mixing is not expected to affect the whole areas. It shall be noticed, however, that oversegmentation appears as a general tendency which seems to feature the proposed method, since the number of detected segments is, on average, higher than the true one.        

Both choices of the kernel appear appropriate at a first sight and such choice is overall not relevant in the easiest settings. However, the use of a bounded Uniform kernel is riskier as expected, and results show higher variability and high sensitivity to a bad selection of the smoothing amount. Hence, the general preference for unbounded-support kernels is confirmed. }


\begin{table}[tb]
\caption{Simulation results for setting C: each entry displays the Monte Carlo average of the Adjusted Rand Index (ARI) and of the number of detected segments (the true number of segments is 2). Standard deviations are reported in brackets. } \label{settingC}
\centering
\resizebox{.49\textwidth}{!}{%
\begin{tabular}{p{1.8cm}p{1.55cm}p{2cm}p{2cm}p{2cm}p{2cm}}
\hline
&  &  \includegraphics[scale=2]{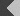} \newline C1: benchmark   & \includegraphics[scale=2]{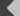} \newline C2: shaded contours  & \includegraphics[scale=2]{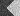} \newline C3: heterogeneous colors   \\ 
\hline\hline
&ARI & 1.00 & 1.00 & 0.80 \\ 
  Normal K & & (0.00) & (0.01) & (0.20) \\ 
$h=h_N$&  \# segments & 2.00 & 2.00 & 3.87 \\ 
  & & (0.00) & (0.00) & (2.14) \\  \hline
&ARI & 1.00 & 0.94 & 0.57 \\ 
  Normal K & & (0.00) & (0.18) & (0.17) \\ 
$h=0.75h_N$&  \# segments & 2.00 & 2.19 & 6.21 \\ 
  & & (0.00) & (0.78) & (2.02) \\  \hline
&ARI & 1.00 & 1.00 & 0.67 \\ 
  Normal K & & (0.00) & (0.01) & (0.20) \\ 
$h=1.25h_N$&  \# segments & 2.00 & 2.00 & 5.15 \\ 
  & & (0.00) & (0.04) & (2.18) \\  \hline\hline
&ARI & 1.00 & 0.98 & 0.94 \\ 
  Uniform K & & (0.00) & (0.07) & (0.11) \\ 
$h=h_N$&  \# segments & 2.00 & 2.13 & 2.54 \\ 
  & & (0.00) & (0.58) & (1.10) \\  \hline
&ARI & 1.00 & 0.97 & 0.83 \\ 
  Uniform K & & (0.00) & (0.15) & (0.17) \\ 
$h=0.75h_N$&  \# segments & 2.00 & 2.01 & 3.55 \\ 
  & & (0.00) & (0.29) & (1.67) \\  \hline
&ARI & 0.29 & 0.59 & 0.71 \\ 
  Uniform K & & (0.32) & (0.46) & (0.37) \\ 
$h=1.25h_N$&  \# segments & 2.56 & 2.02 & 2.82 \\ 
  & & (1.55) & (1.23) & (1.88) \\   
   \hline \hline
\end{tabular}}
\label{settingC}
\end{table}

\begin{table}[tb]
\caption{Simulation results for setting D: each entry display the Monte Carlo average of the Adjusted Rand Index (ARI) and of the number of detected segments (the true number of segments is 3). Standard deviations are reported in brackets. } \label{settingD}
\centering
\resizebox{.49\textwidth}{!}{%
\begin{tabular}{p{1.8cm}p{1.55cm}p{2cm}p{2cm}p{2cm}}
  \hline
&  &  \includegraphics[scale=2]{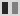} \newline D1: benchmark   & \includegraphics[scale=2]{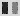} \newline D2: shaded contours  & \includegraphics[scale=2]{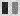} \newline D3: heterogeneous colors   \\ 
  \hline \hline
&ARI & 0.49 & 0.24 & 0.38 \\ 
Normal K & & (0.09) & (0.25) & (0.07) \\ 
$h=h_N$&  \# segments & 7.16 & 2.77 & 9.28 \\ 
  & & (1.47) & (2.01) & (1.85) \\  \hline
&ARI & 0.42 & 0.12 & 0.36 \\ 
  Normal K & & (0.09) & (0.19) & (0.06) \\ 
 $h=0.75h_N$&  \# segments & 8.94 & 2.55 & 10.59 \\ 
  & & (1.69) & (2.28) & (1.72) \\  \hline
&ARI & 0.61 & 0.13 & 0.39 \\ 
  Normal K & & (0.34) & (0.20) & (0.12) \\ 
  $h=1.25h_N$&  \# segments & 3.12 & 2.93 & 7.97 \\ 
  & & (1.97) & (2.24) & (2.01) \\ \hline\hline
&ARI & 1.00 & 0.28 & 0.36 \\ 
  Uniform K & & (0.00) & (0.24) & (0.07) \\ 
  $h=h_N$&  \# segments & 3.00 & 3.60 & 9.87 \\ 
  & & (0.00) & (2.53) & (1.90) \\  \hline
&ARI & 1.00 & 0.42 & 0.42 \\ 
  Uniform K & & (0.00) & (0.17) & (0.13) \\ 
  $h=0.75h_N$&  \# segments & 3.00 & 3.70 & 7.53 \\ 
  & & (0.00) & (2.22) & (1.89) \\  \hline
&ARI & 0.81 & 0.39 & 0.36 \\ 
  Uniform K & & (0.26) & (0.30) & (0.06) \\ 
  $h=1.25h_N$&  \# segments & 3.63 & 3.72 & 10.37 \\ 
  & & (1.88) & (2.31) & (1.85) \\ 
  \hline\hline
\end{tabular}}
\label{settingD}
\end{table}

\subsection{Real images illustration}
In this section, some examples are presented to illustrate the proposed procedure in action, and to evaluate its effectiveness in identifying the salient regions for varying characteristics of more challenging 
images. Further examples are presented in the online Supplementary Material. Grayscale and multichannel images are considered, and selected for featuring either shaded and neat colors, and possible highlighted contours; 
both convex and nonconvex shaped segments have been analyzed. The procedure is here compared with the performance  of some competitors: as a benchmark methods, $K-$means clustering is considered, thresholding 
based on the Otsu algorithm, 
and an edge-detection algorithm, based on the Sobel filter \cite[see, for details,][\S 10.2.1 and, respectively, \S 4.1]{Dougharty09}. 
The first method has been given a head start by setting the number of clusters to its true number, as it is intuitively perceived by the author. The choice is not always obvious, especially for photos or, in general, shaded-color images.  
The second algorithm assumes that the image contains two classes of pixels -black/white- grossly corresponding to two modal regions of the histogram built on the gray intensities. It calculates the optimum threshold separating the two classes based on the minimization of 
the intra-class variance. Although it is designed for grayscale images only, thus requiring a prior preprocessing of multi-channel images, it has been considered for comparison as it 
represents a rough, yet effective, binary variant of the proposed procedure. 
The third method works in a slightly different logic, as it is aimed at edge rather than segment detection. Then, segments are assumed to be the sets of pixels within the linked edges. 

A further aim of the empirical work is to investigate the use of the density function as a tool to identify the main features of an image: one question of interest is its ability to detect edges of the segments. Also, in agreement with the previous section, its sensitivity to different smoothing amounts is evaluated, {by testing $h \in \{h_N, 0.75, h_N, 1.25h_N\}$. Due to the considerations arisen in the simulation study, a Gaussian kernel only has been considered here.} 


Additionally, the ability of the cluster tree to identify a hierarchy of cluster merging that is meaningful with respect to the image has been tested. Finally, the allocation of low-density pixels, not belonging to the segment cores has been observed and commented.

\begin{figure*}[tb]
\begin{center}
\begin{tabular}{p{3.1cm}p{3.1cm}p{3.1cm}p{3.1cm}p{3.1cm}}
\fontsize{7}{10}\selectfont {original image} & \fontsize{7}{10}\selectfont {$K-$means segmentation, $K = 5$} & \fontsize{7}{10}\selectfont {thresholding segmentation} & \fontsize{7}{10}\selectfont {Sobel segmentation} & \fontsize{7}{10}\selectfont {nonparametric segmentation, $h = h_N$}\\
\vspace{-0cm}\includegraphics[width=.175\textwidth,height=1.6cm]{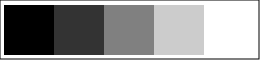}&
\includegraphics[height=.175\textwidth, width=1.6cm,angle=-90]{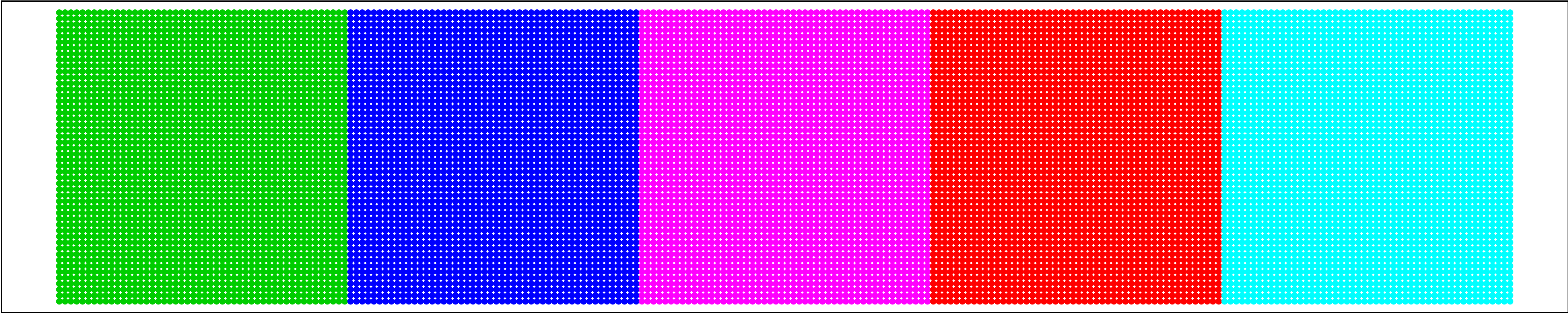}&
\includegraphics[width=.175\textwidth, height= 1.6cm,angle= -180]{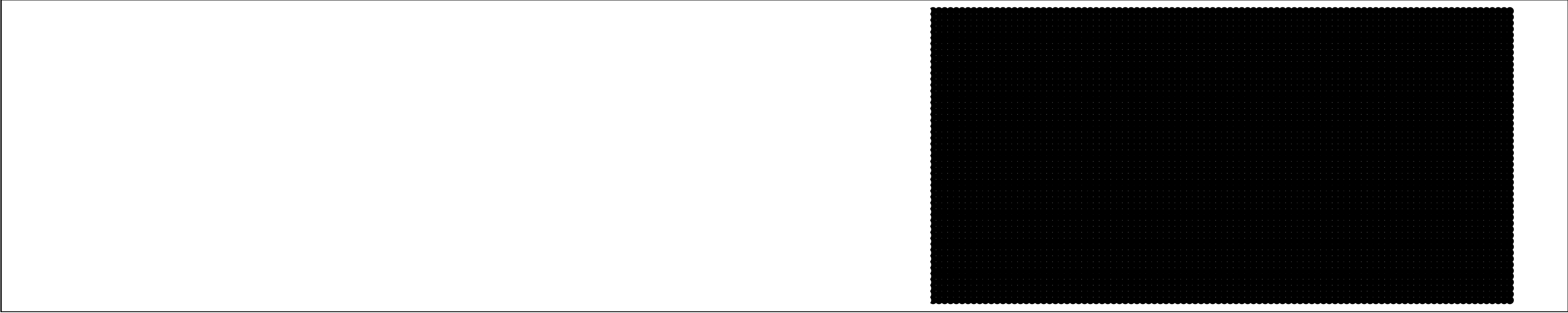}&
\raisebox{-\height}{\includegraphics[width=.175\textwidth, height=1.6cm]{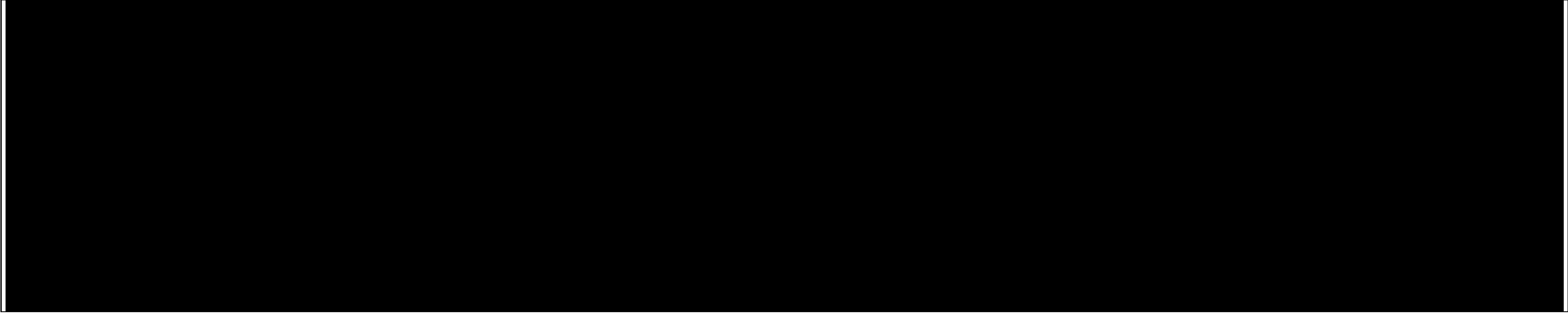}}&
\includegraphics[height=.175\textwidth, width=1.6cm,angle=-90]{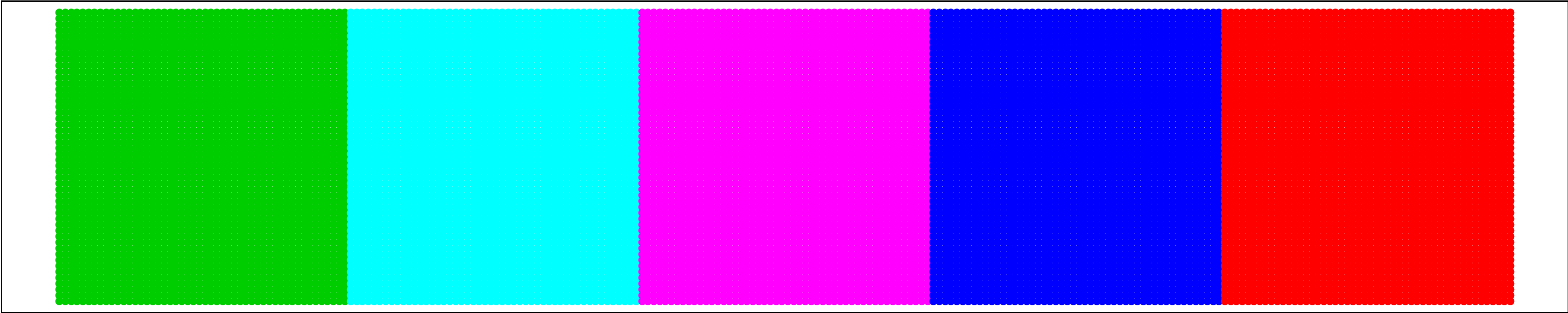}\\
\fontsize{7}{10}\selectfont { segment cores, $h = h_N$} & \fontsize{7}{10}\selectfont { cluster tree, $h = h_N$} & \fontsize{7}{10}\selectfont { density contours, $h = h_N$} & \fontsize{7}{10}\selectfont { nonparametric segmentation, $h = 0.75h_N$} & \fontsize{7}{10}\selectfont { nonparametric segmentation, $h = 1.25h_N$}\\
&&&&\\
\vspace{-2cm}\includegraphics[height=.175\textwidth, width=1.6cm, angle=-90]{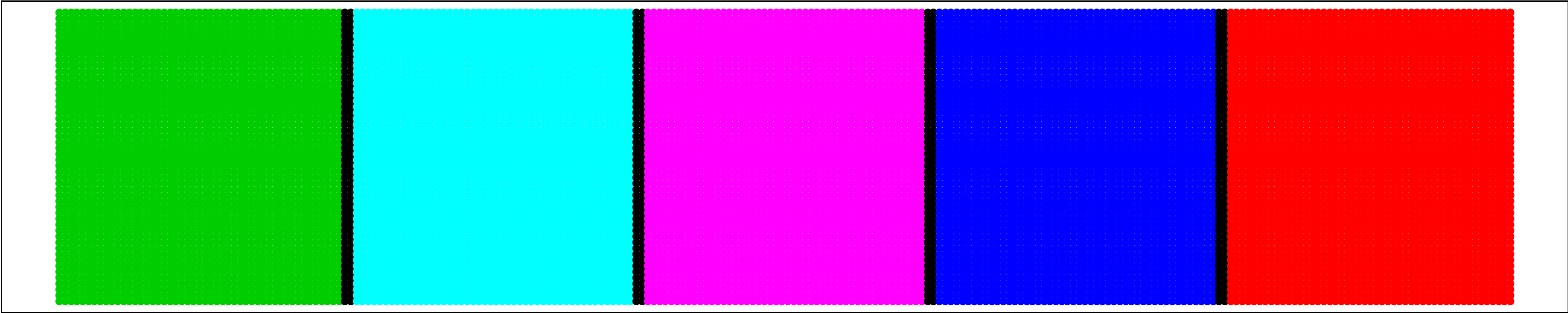}
&
\includegraphics[height=1.6cm, width=.175\textwidth]{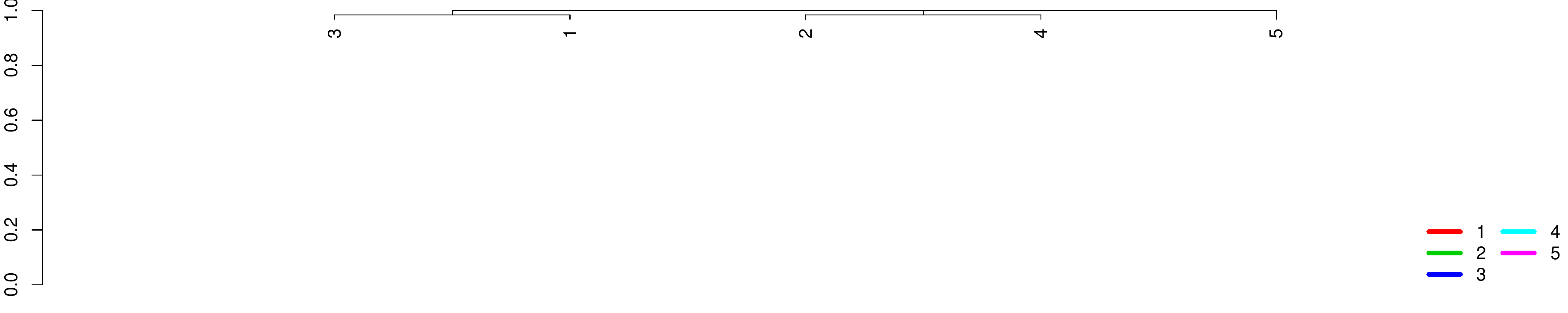}&
\includegraphics[height=1.6cm, width=.175\textwidth]{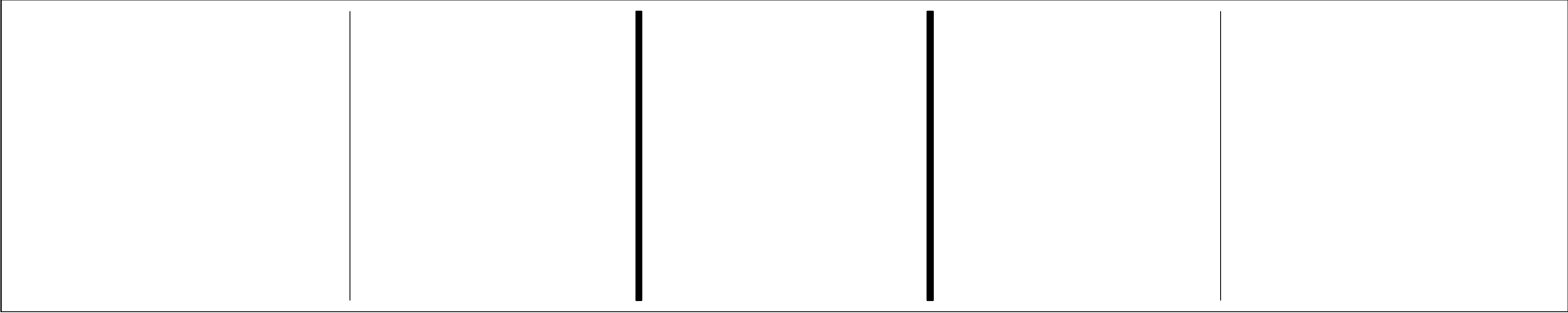}&
\vspace{-1.95cm}\includegraphics[width=1.6cm, height=.175\textwidth, angle=-90]{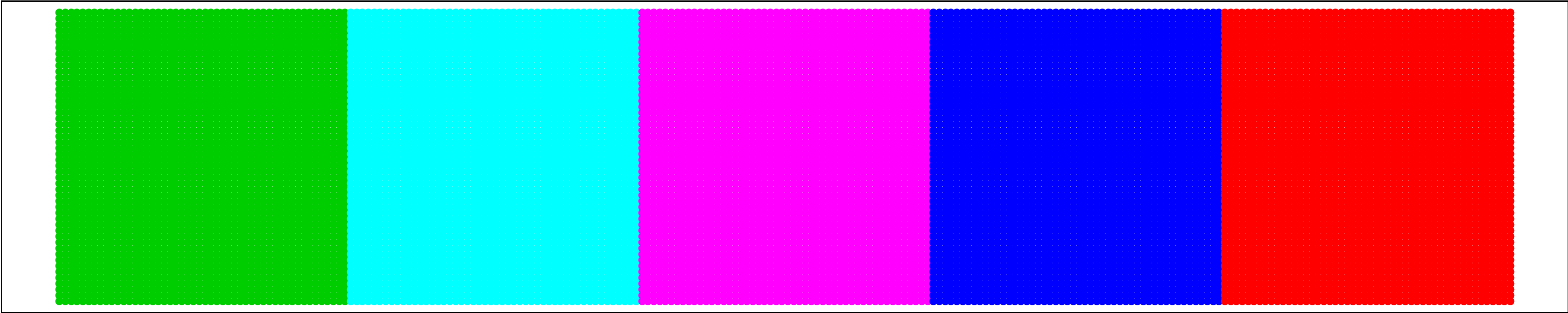}&
\vspace{-1.95cm}\includegraphics[width=1.6cm, height=.175\textwidth, angle=-90]{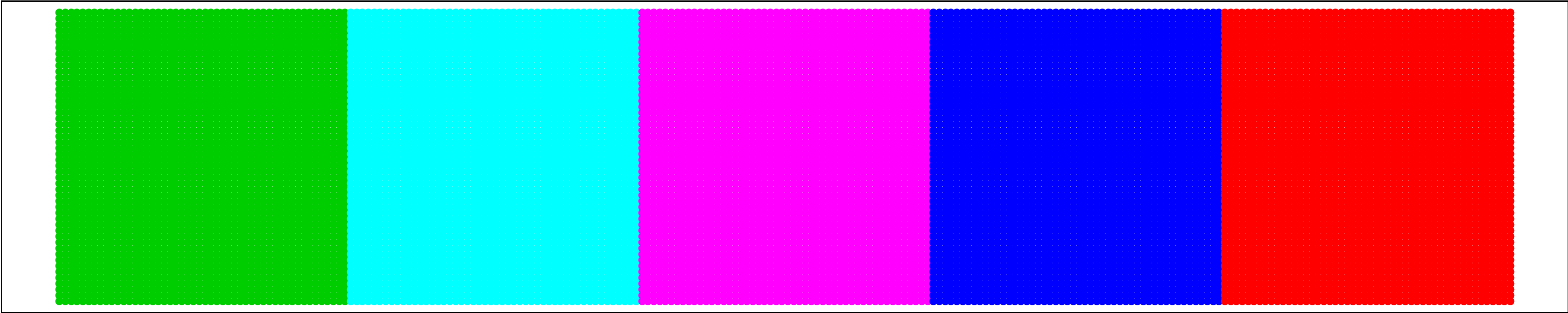}
\end{tabular}
\end{center}
\caption{Segmentation results. Segments have been assigned arbitrary colors, except for the thresholding segmentation, where segments are either black or white by construction.}\label{fig:bnsquare}
\end{figure*}

\begin{figure*}[tb]
\begin{center}
\begin{tabular}{p{3.1cm}p{3.3cm}p{3.1cm}p{3.1cm}p{3.1cm}}
\fontsize{7}{10}\selectfont {original image} & \fontsize{7}{10}\selectfont {$K-$means segmentation, $K = 14$} & \fontsize{7}{10}\selectfont {thresholding segmentation} & \fontsize{7}{10}\selectfont {Sobel segmentation} & \fontsize{7}{10}\selectfont {nonparametric segmentation, $h = h_N$}\\
\vspace{-0cm}\includegraphics[width=.175\textwidth,height=4cm]{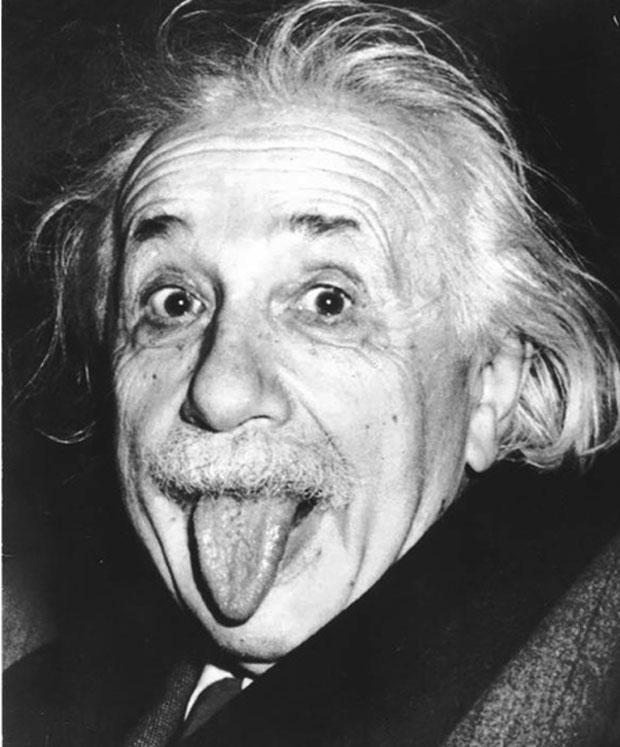}&
\vspace{-0.3cm}\includegraphics[height=.18\textwidth, width=4cm,angle=-90]{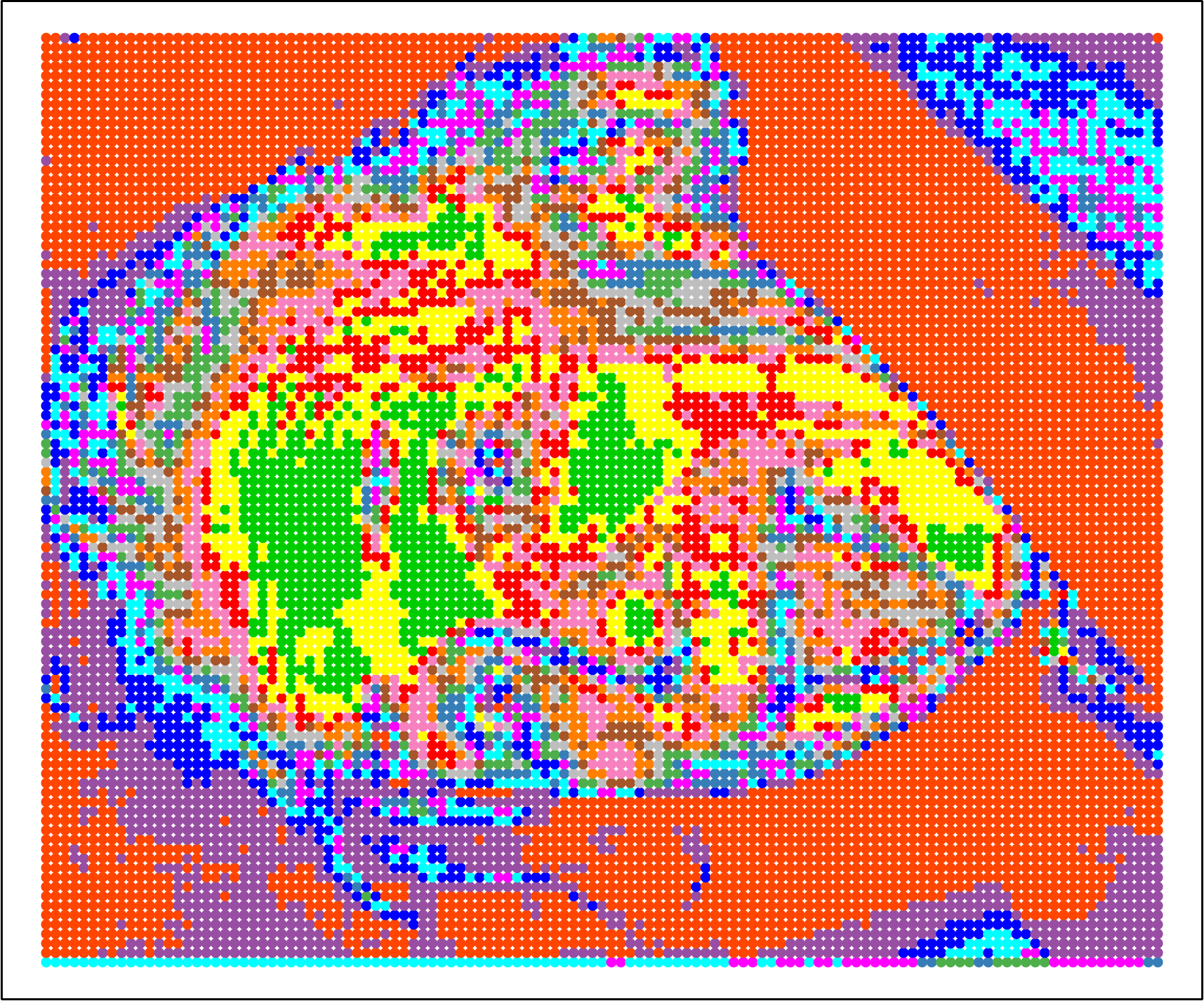}&
\includegraphics[width=.175\textwidth, height= 4cm,angle= -180]{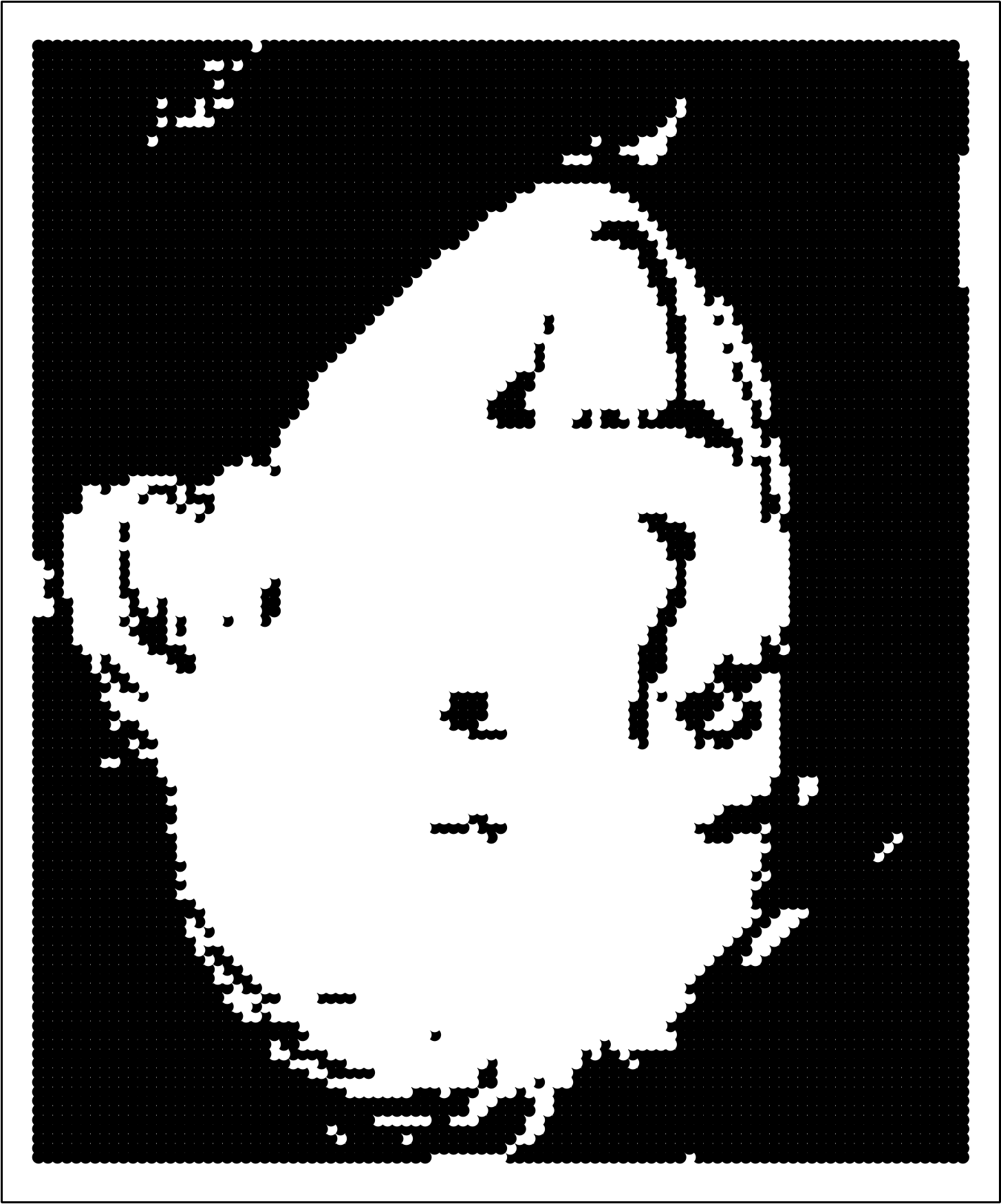}&
\raisebox{-\height}{\includegraphics[width=.175\textwidth, height=4cm]{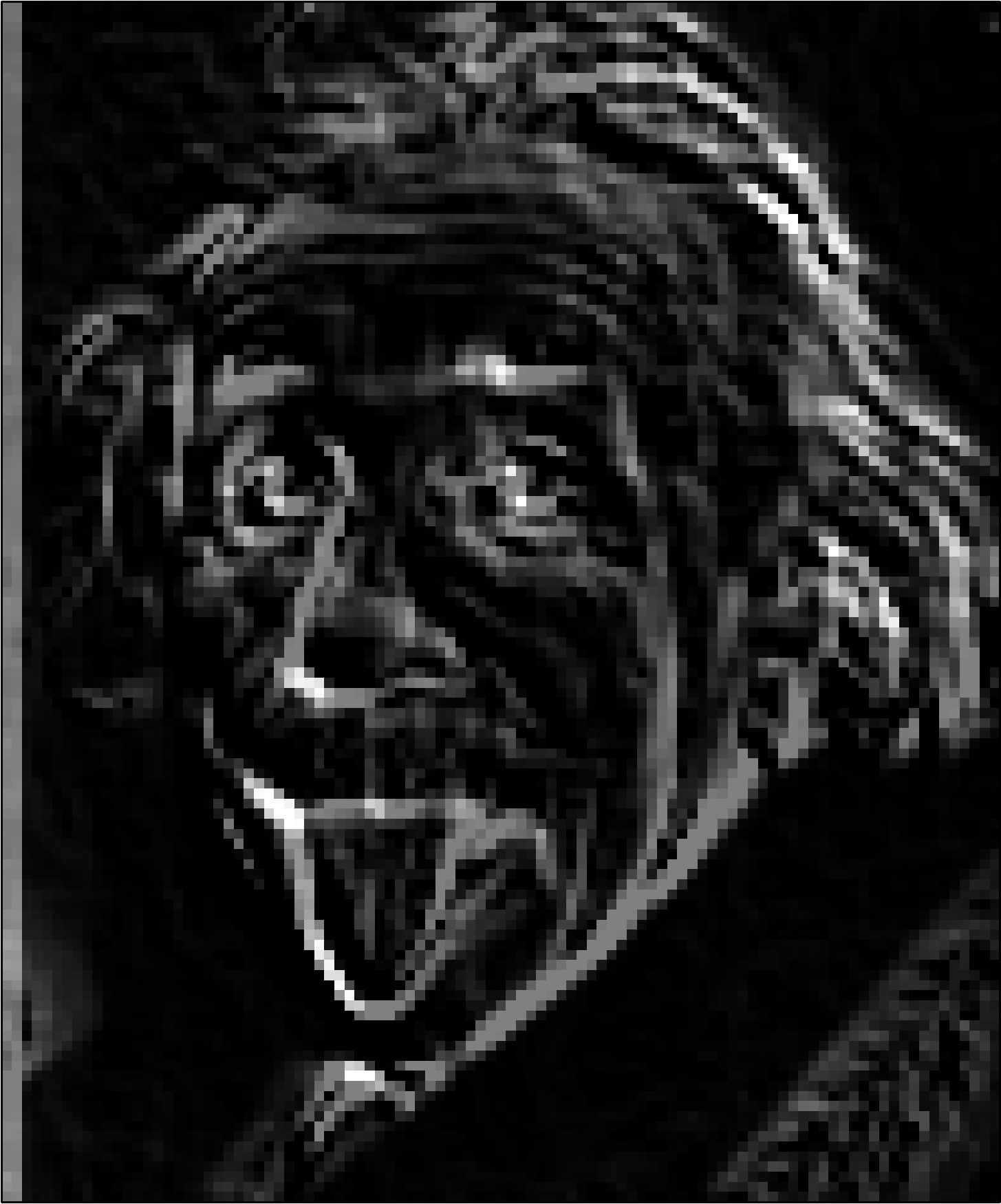}}&
\includegraphics[height=.175\textwidth, width=4cm,angle=-90]{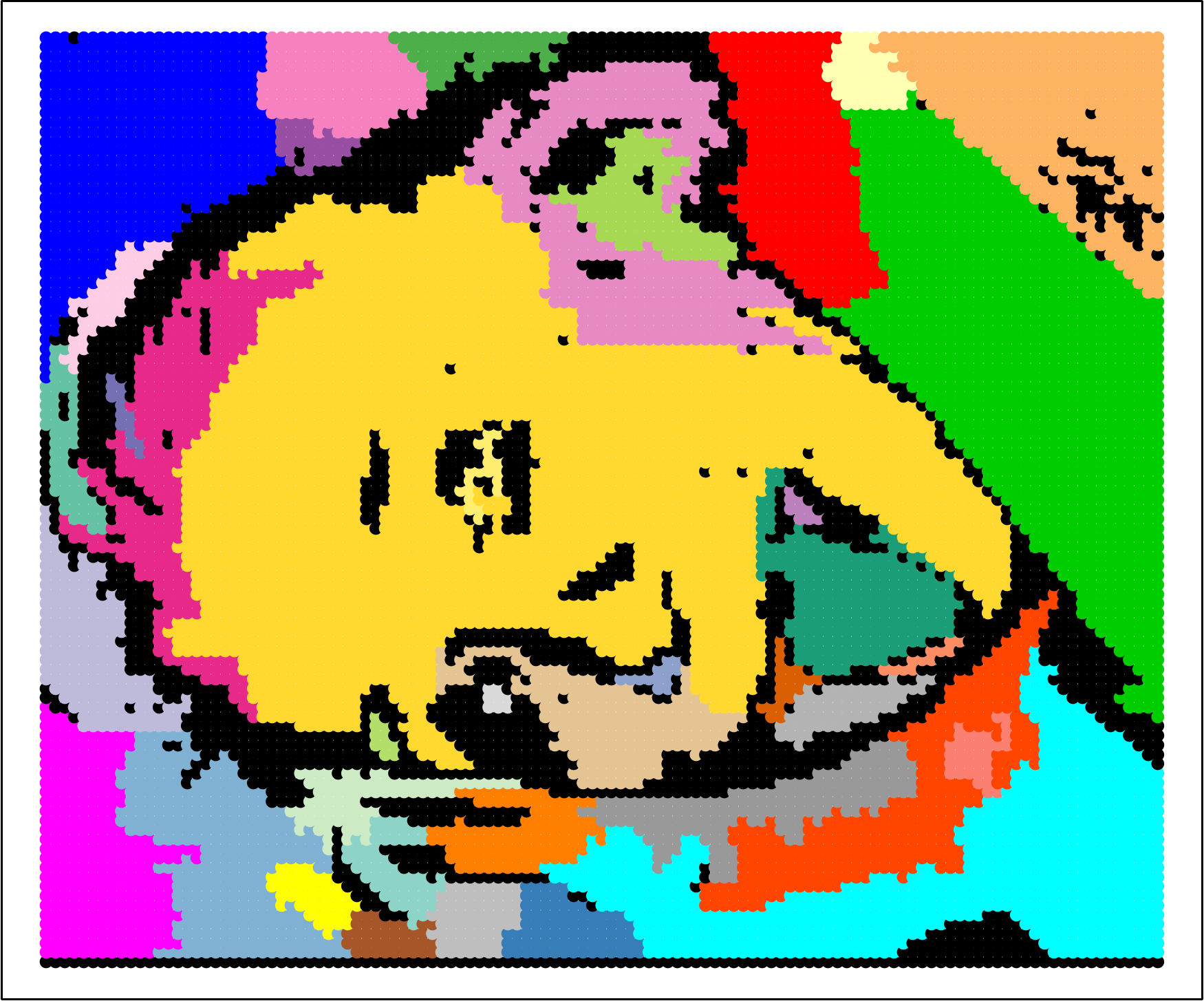}\\
\fontsize{7}{10}\selectfont { segment cores, $h = h_N$} & \fontsize{7}{10}\selectfont { cluster tree, $h = h_N$} & \fontsize{7}{10}\selectfont { density contours, $h = h_N$} & \fontsize{7}{10}\selectfont { nonparametric segmentation, $h = 0.75h_N$} & \fontsize{7}{10}\selectfont { nonparametric segmentation, $h = 1.25h_N$}\\
&&&&\\
\vspace{-4.2cm}\includegraphics[height=.175\textwidth, width=4cm, angle=-90]{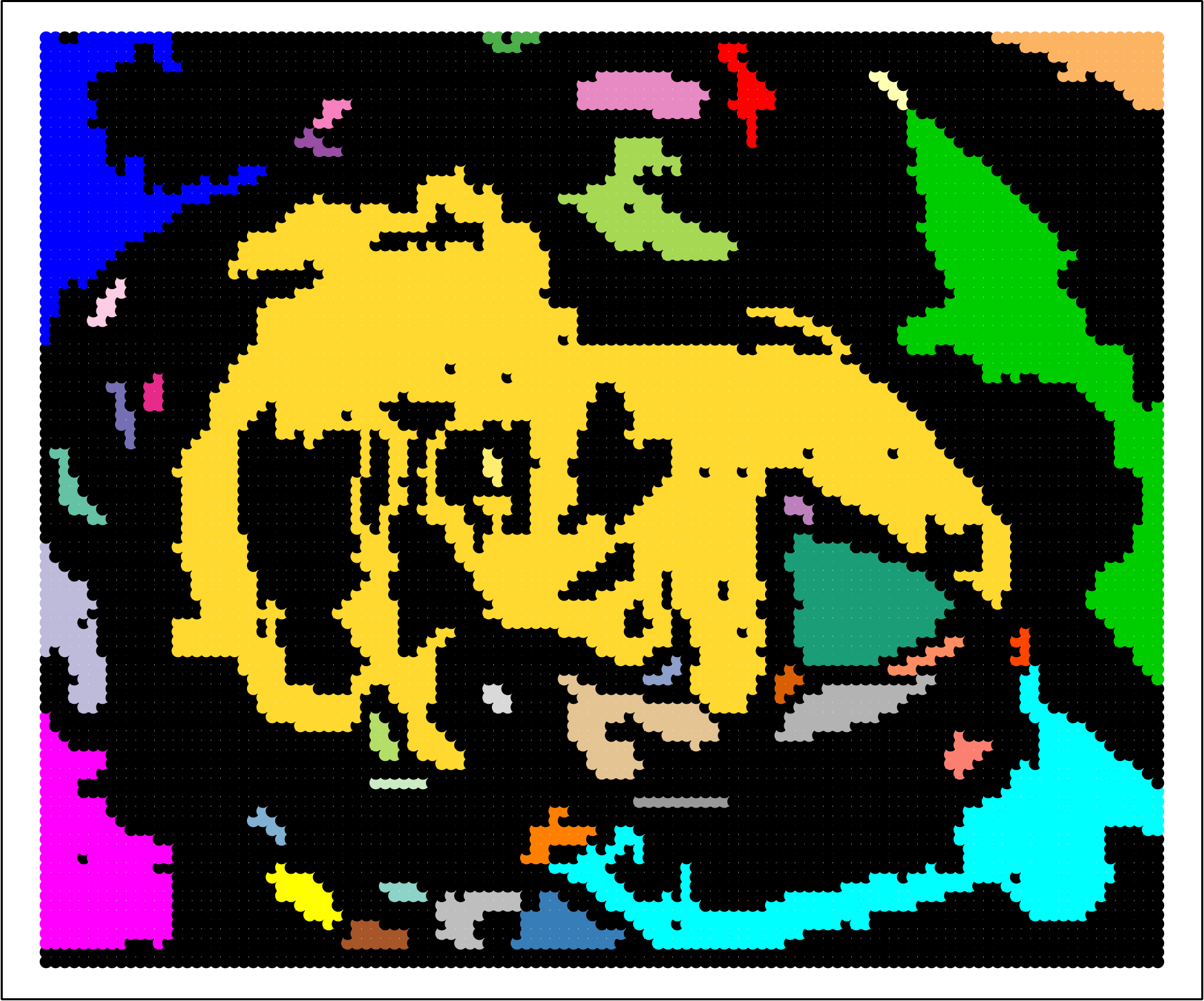}
&
\includegraphics[height=4cm, width=.175\textwidth]{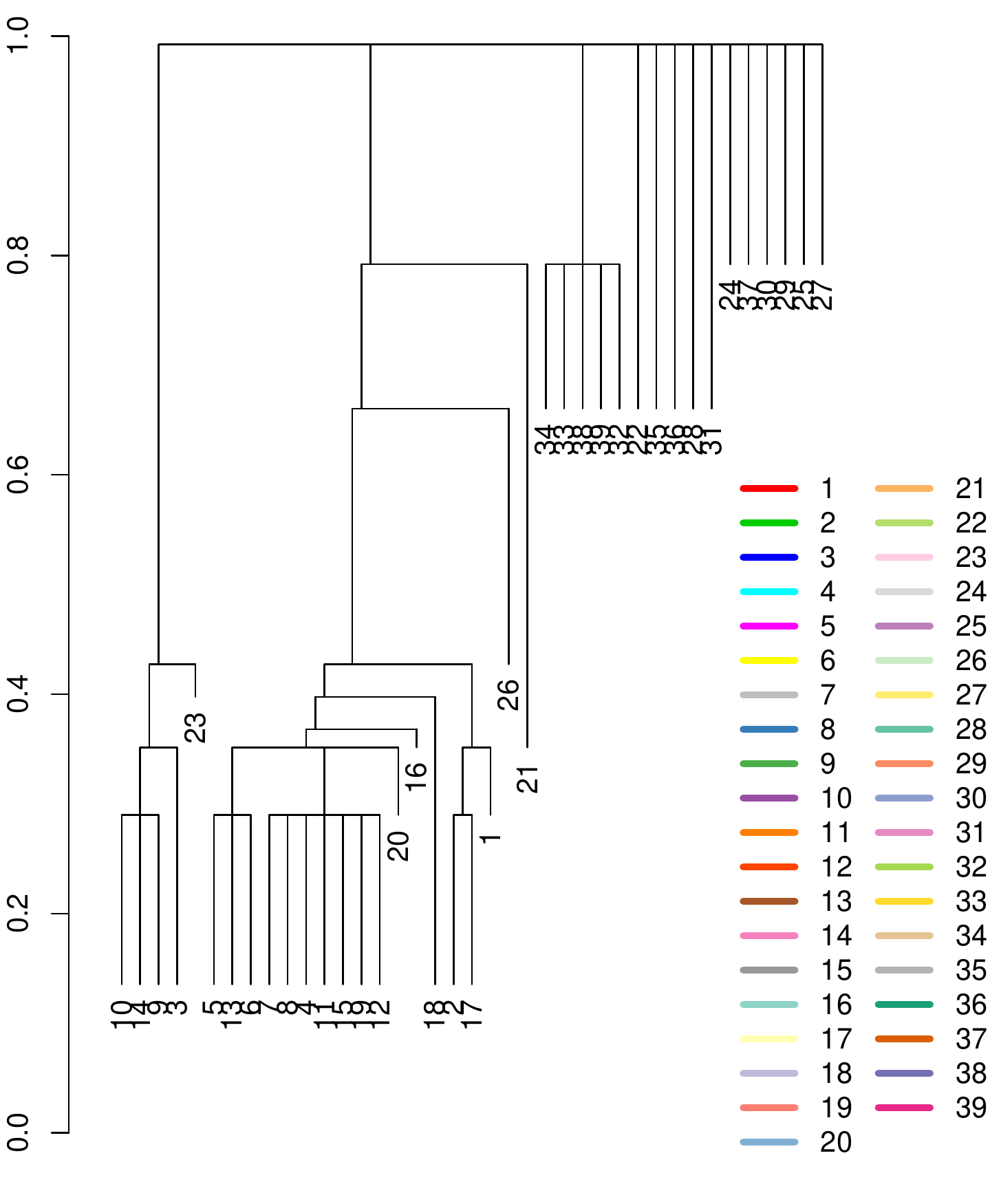}&
\includegraphics[height=4cm, width=.175\textwidth]{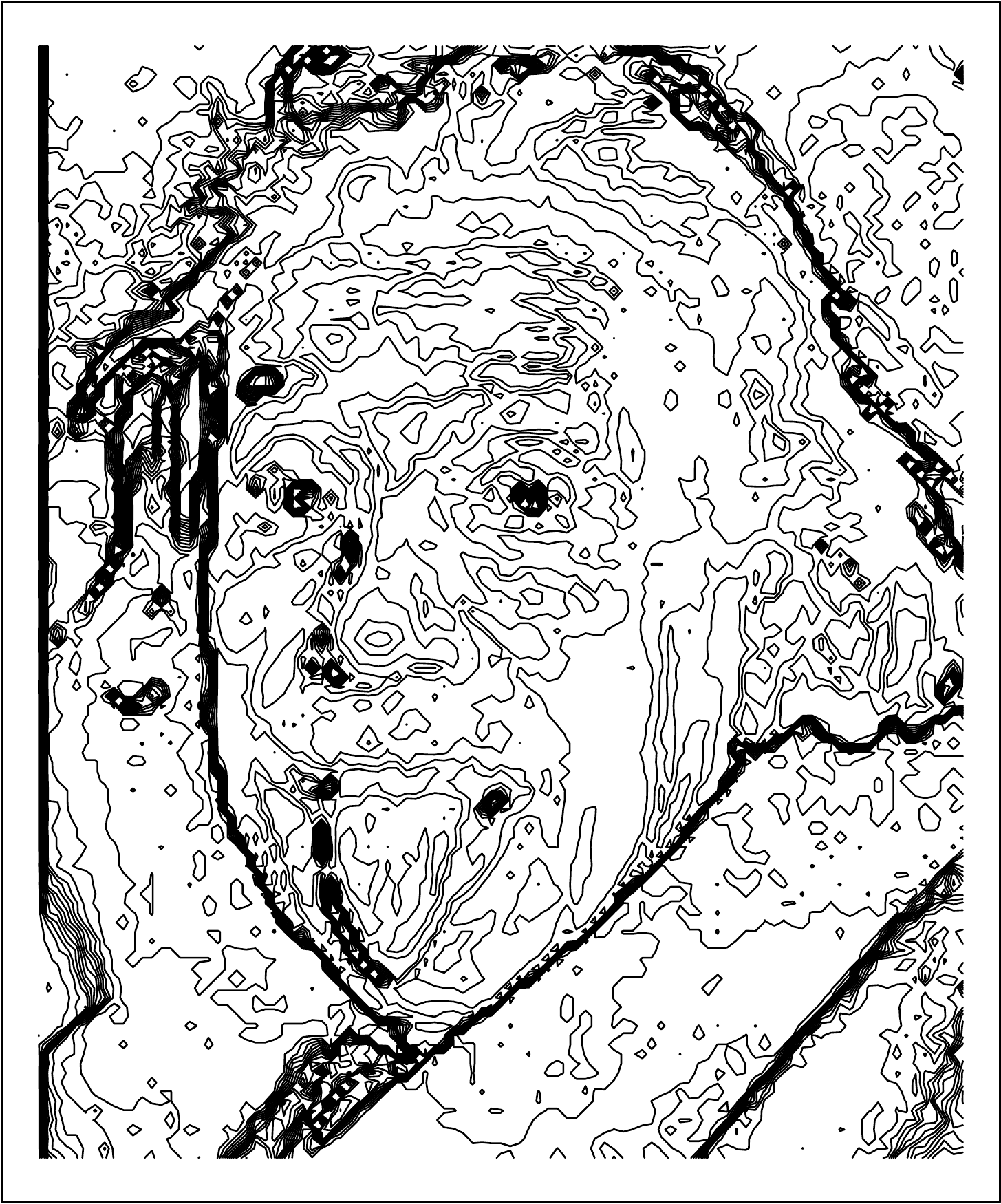}&
\vspace{-4.2cm}\includegraphics[height=.175\textwidth, width=4cm, angle=-90]{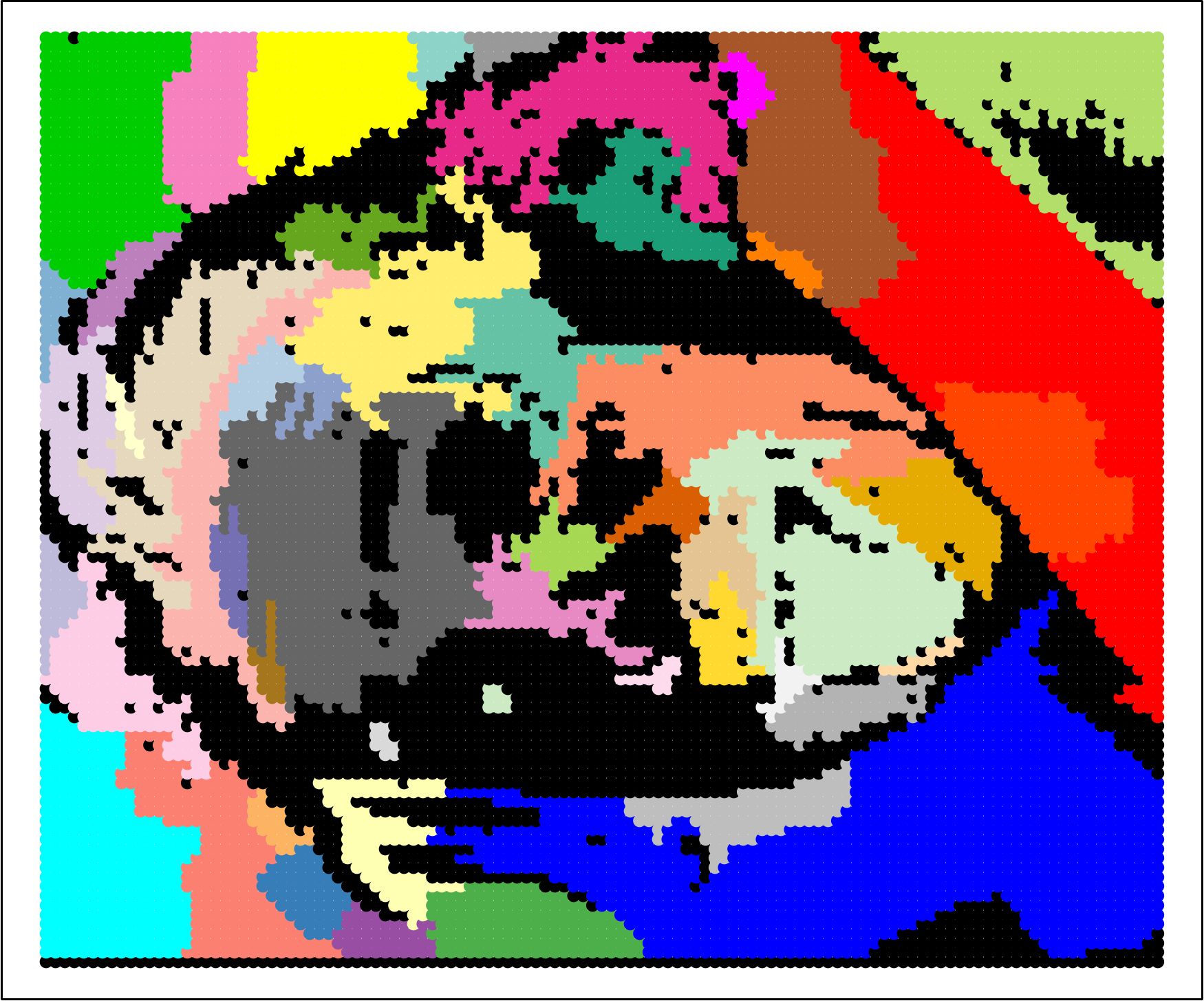}
&
\vspace{-4.2cm}\includegraphics[height=.175\textwidth, width=4cm, angle=-90]{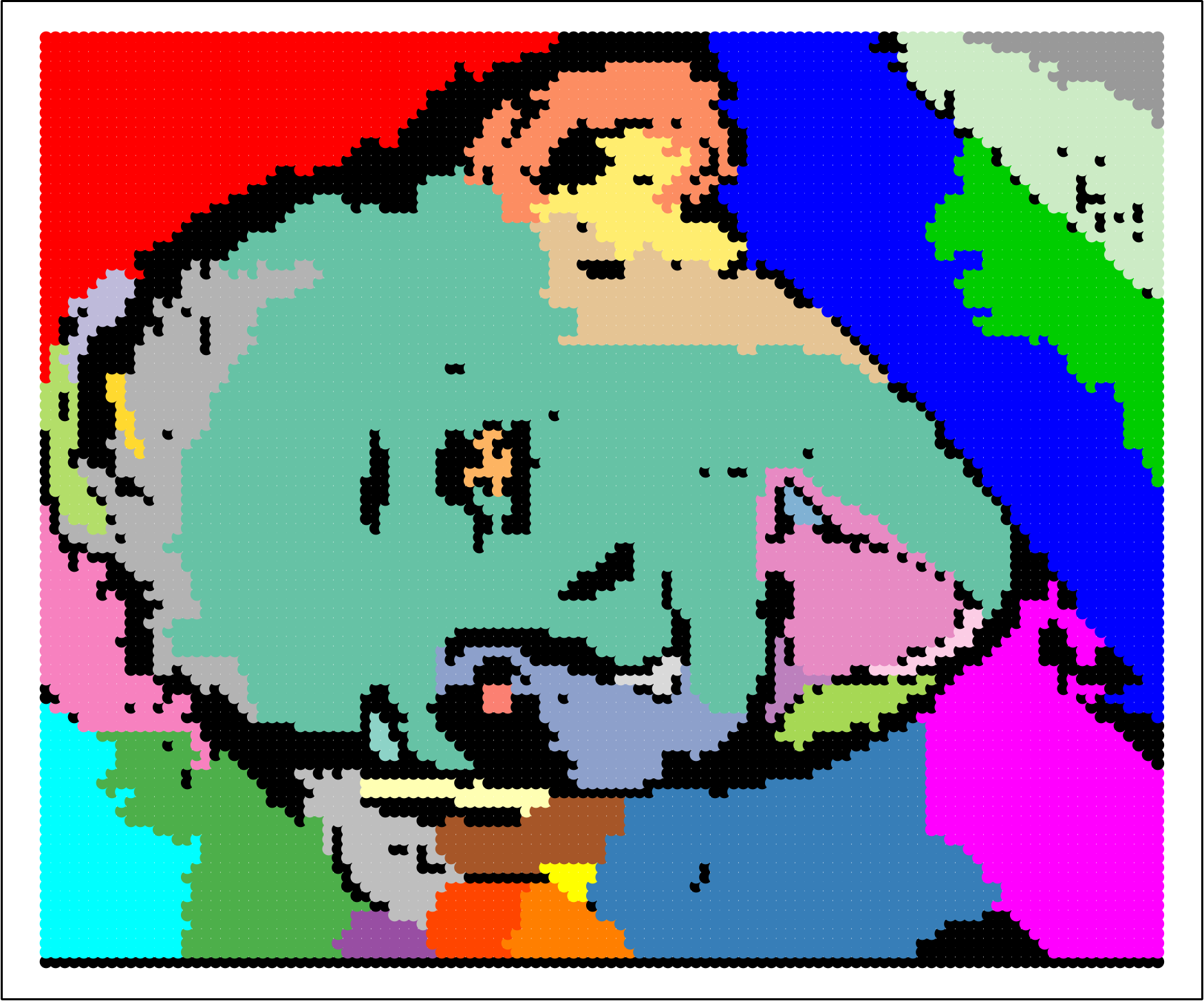}
\end{tabular}
\end{center}
\caption{Segmentation results. Segments have been assigned arbitrary colors, except for the thresholding segmentation, where segments are either black or white by construction.}\label{fig:ein} 
\end{figure*}

\begin{figure*}[tb]
\begin{center}
\begin{tabular}{p{3.1cm}p{3.1cm}p{3.1cm}p{3.1cm}p{3.1cm}}
\fontsize{7}{10}\selectfont {original image} & \fontsize{7}{10}\selectfont {$K-$means segmentation, $K = 5$} & \fontsize{7}{10}\selectfont {thresholding segmentation} & \fontsize{7}{10}\selectfont {Sobel segmentation} & \fontsize{7}{10}\selectfont {nonparametric segmentation, $h = h_N$}\\
\vspace{-0cm}\includegraphics[width=.175\textwidth,height=4cm]{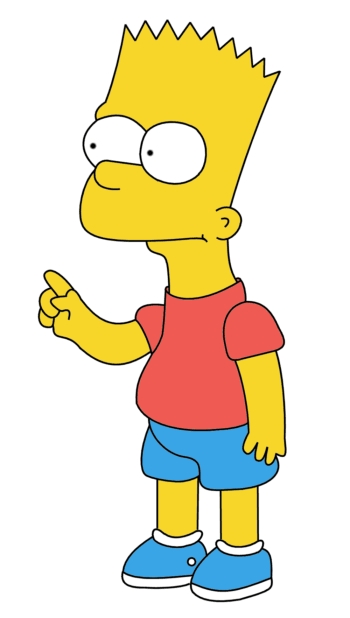}&
\includegraphics[height=.18\textwidth, width=4cm,angle=-90]{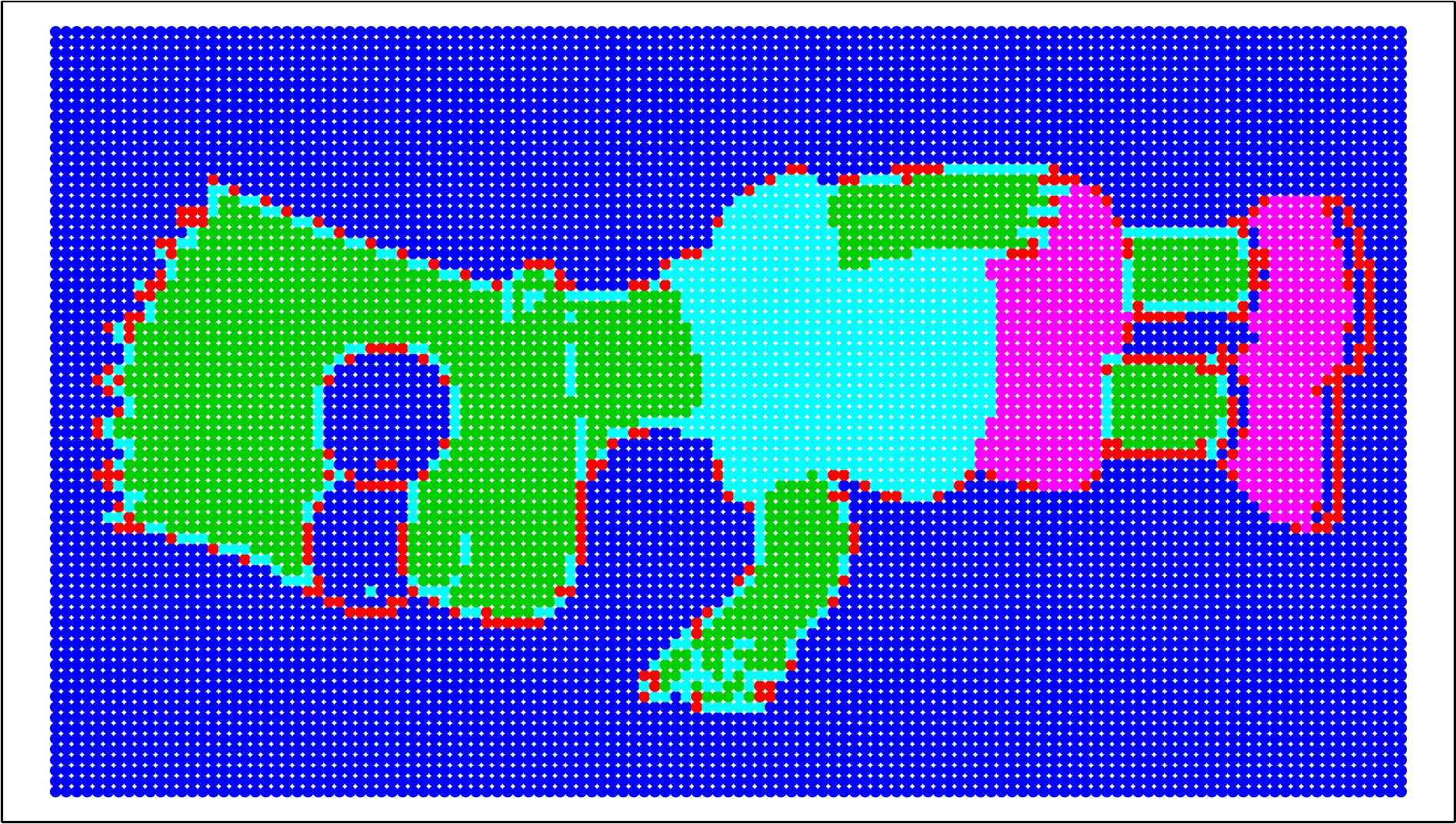}&
\includegraphics[width=.175\textwidth, height= 4cm,angle= -180]{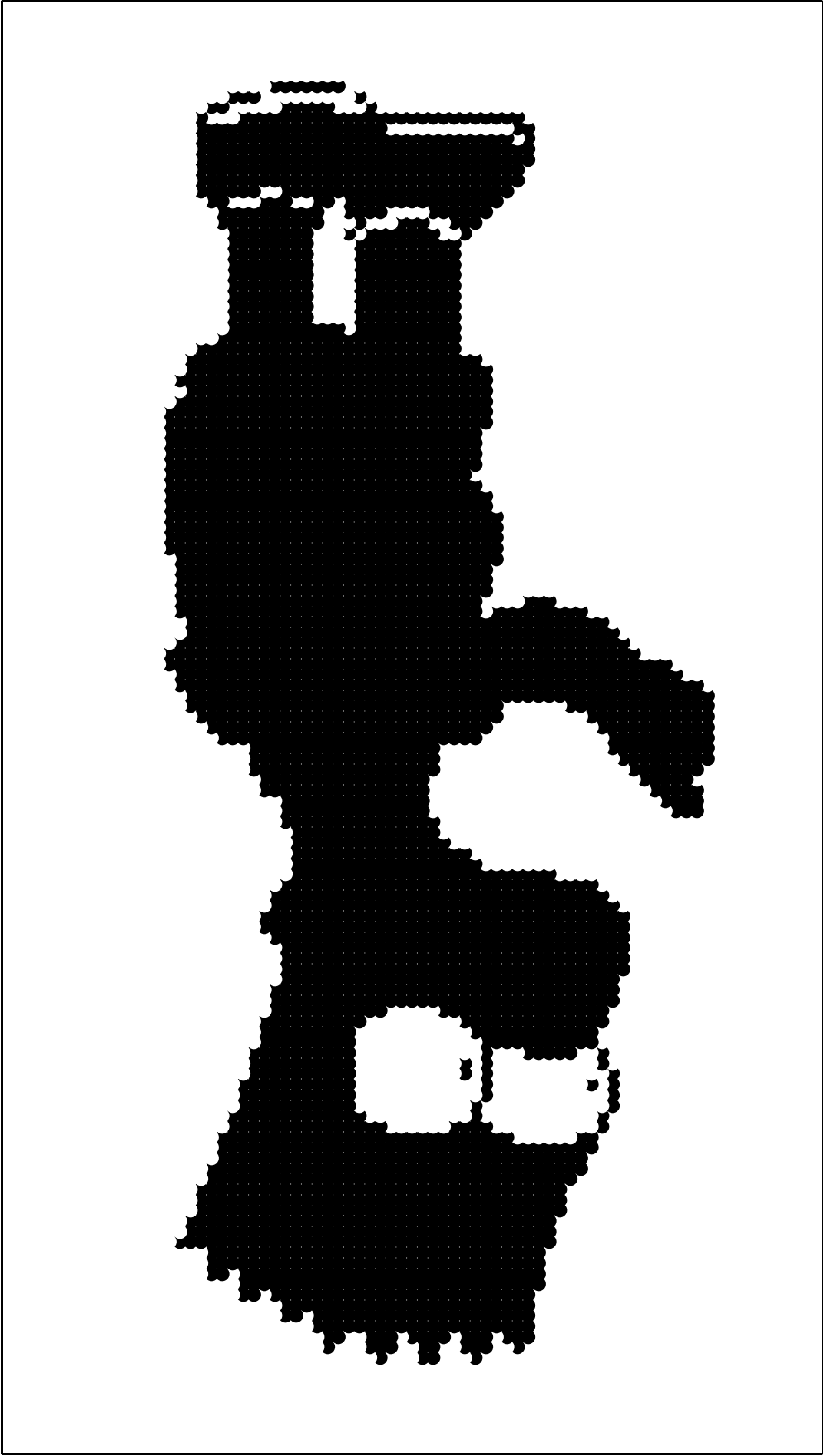}&
\raisebox{-\height}{\includegraphics[width=.175\textwidth, height=4cm]{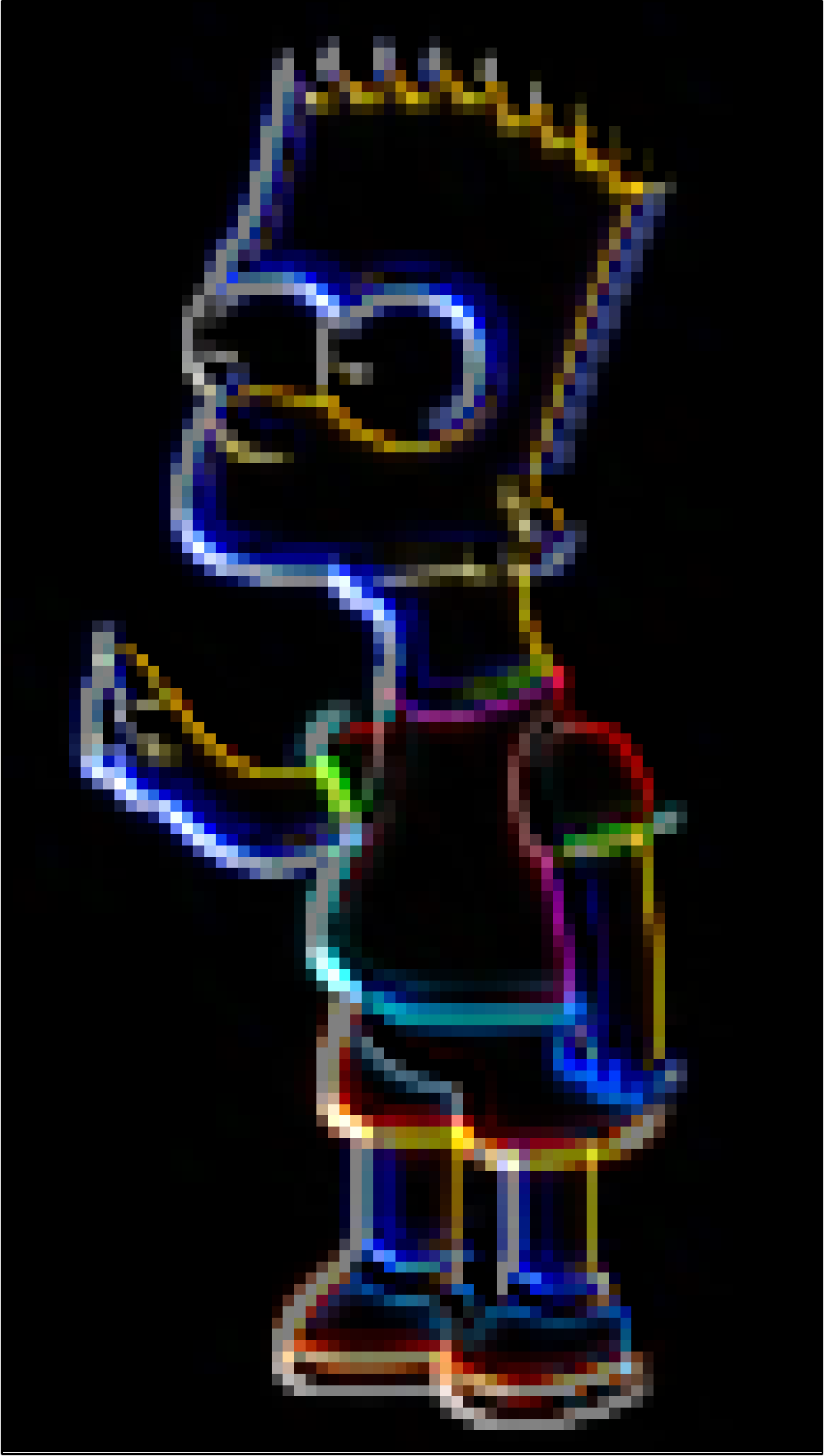}}&
\includegraphics[height=.175\textwidth, width=4cm,angle=-90]{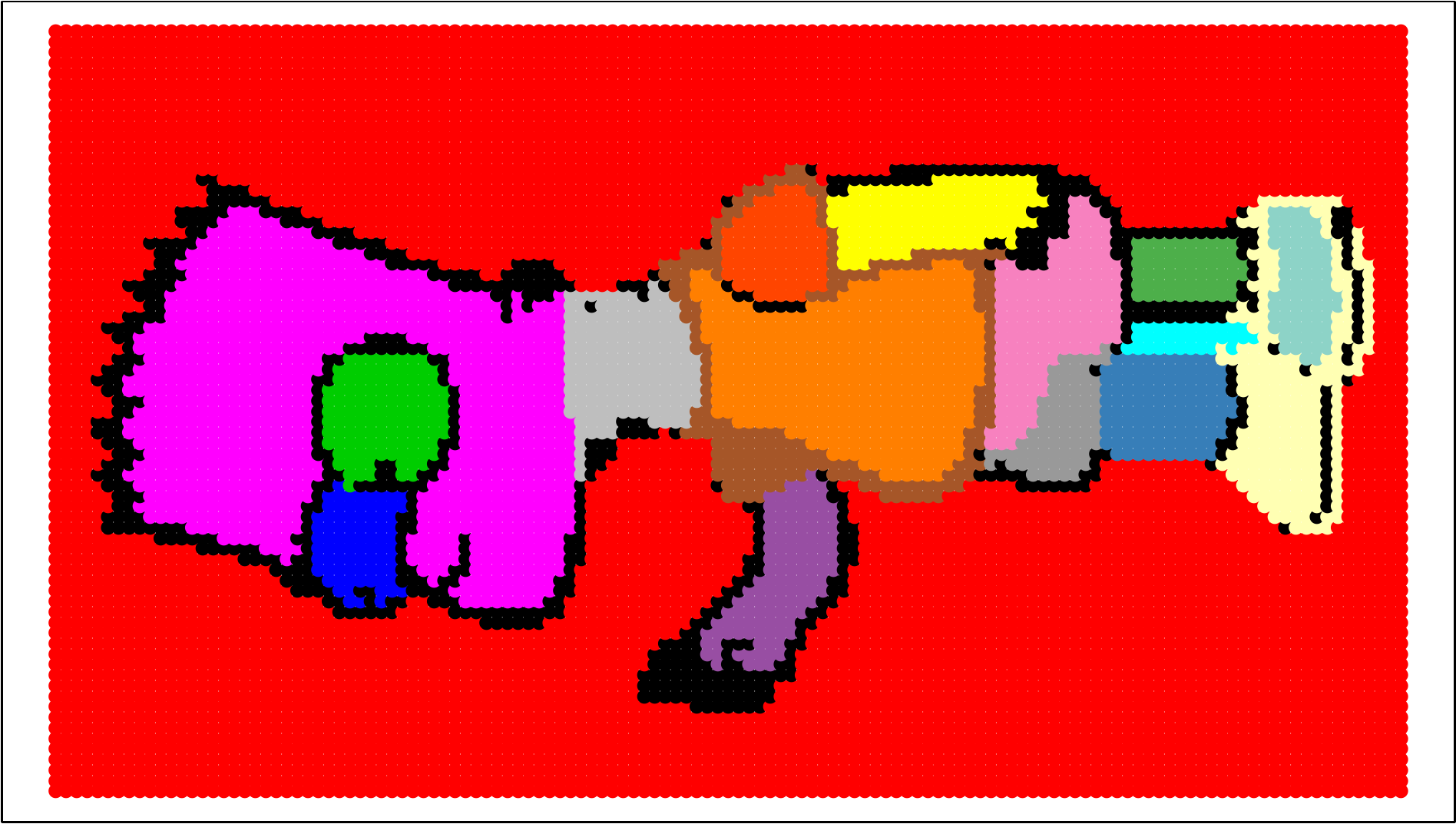}\\
\fontsize{7}{10}\selectfont { segment cores, $h = h_N$} & \fontsize{7}{10}\selectfont { cluster tree, $h = h_N$} & \fontsize{7}{10}\selectfont { density contours, $h = h_N$} & \fontsize{7}{10}\selectfont { nonparametric segmentation, $h = 0.75h_N$} & \fontsize{7}{10}\selectfont { nonparametric segmentation, $h = 1.25h_N$}\\
&&&&\\
\vspace{-4.25cm}\includegraphics[height=.175\textwidth, width=4cm, angle=-90]{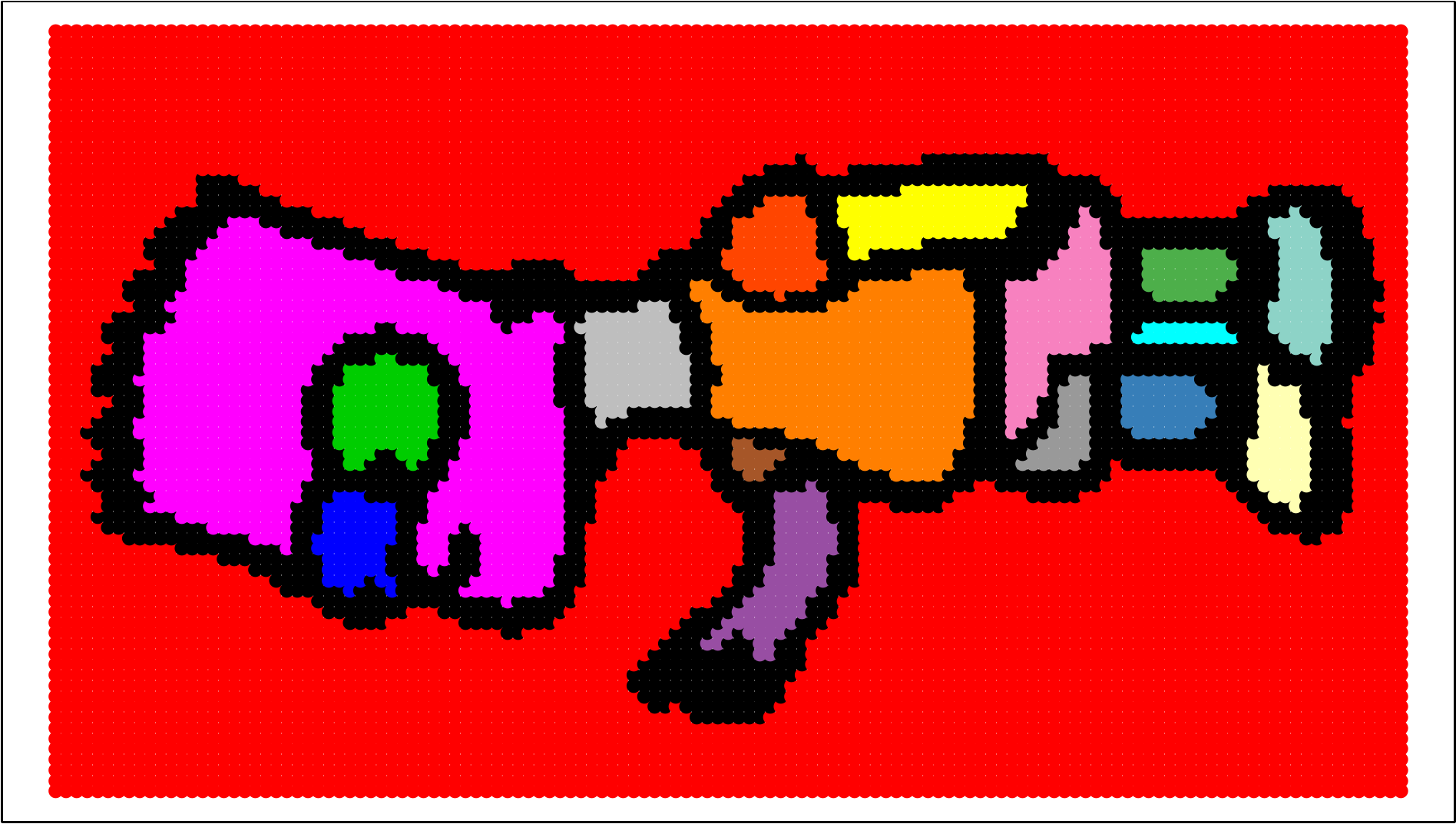}
&
\includegraphics[height=4cm, width=.175\textwidth]{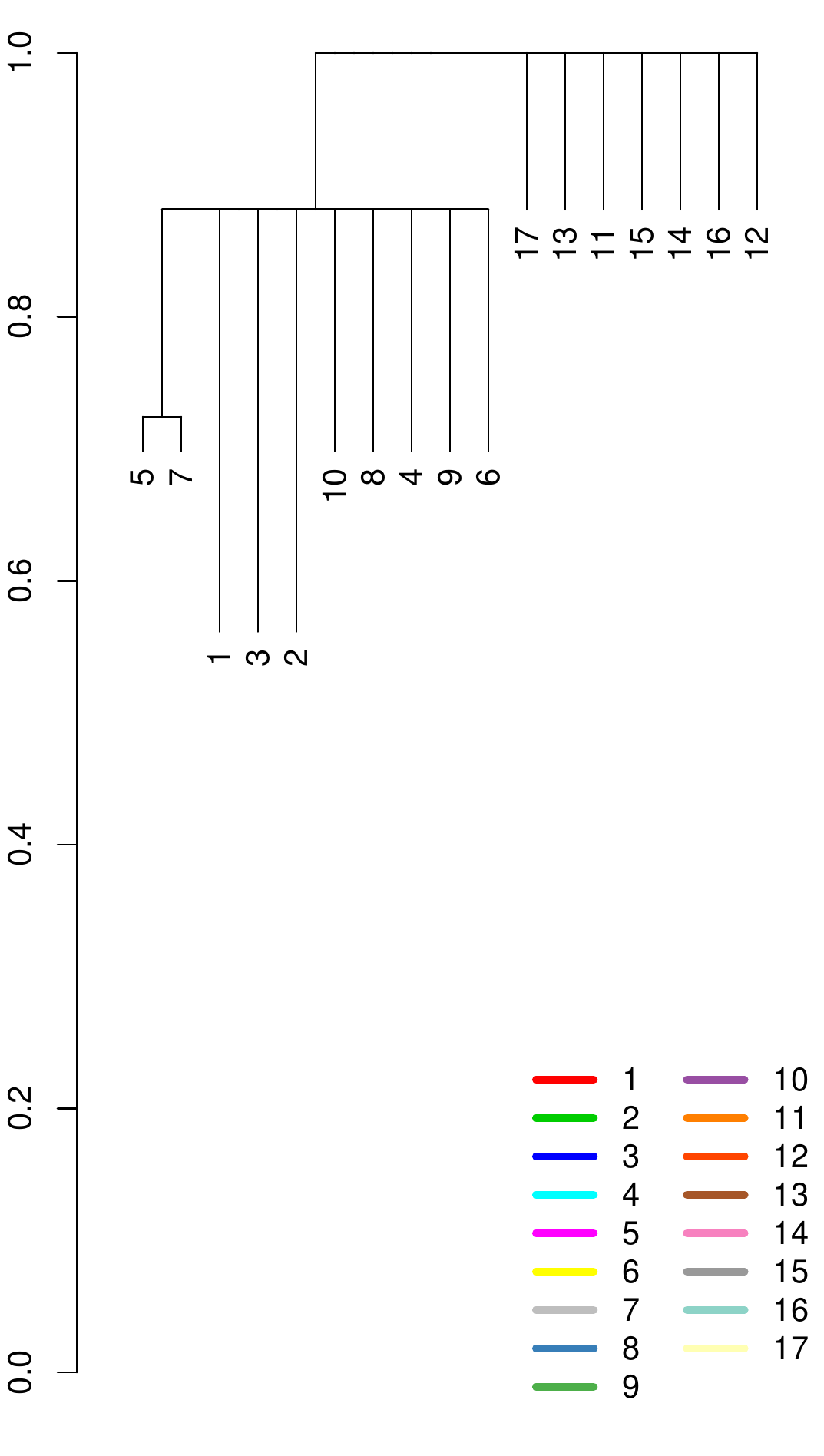}
&
\includegraphics[height=4cm, width=.175\textwidth]{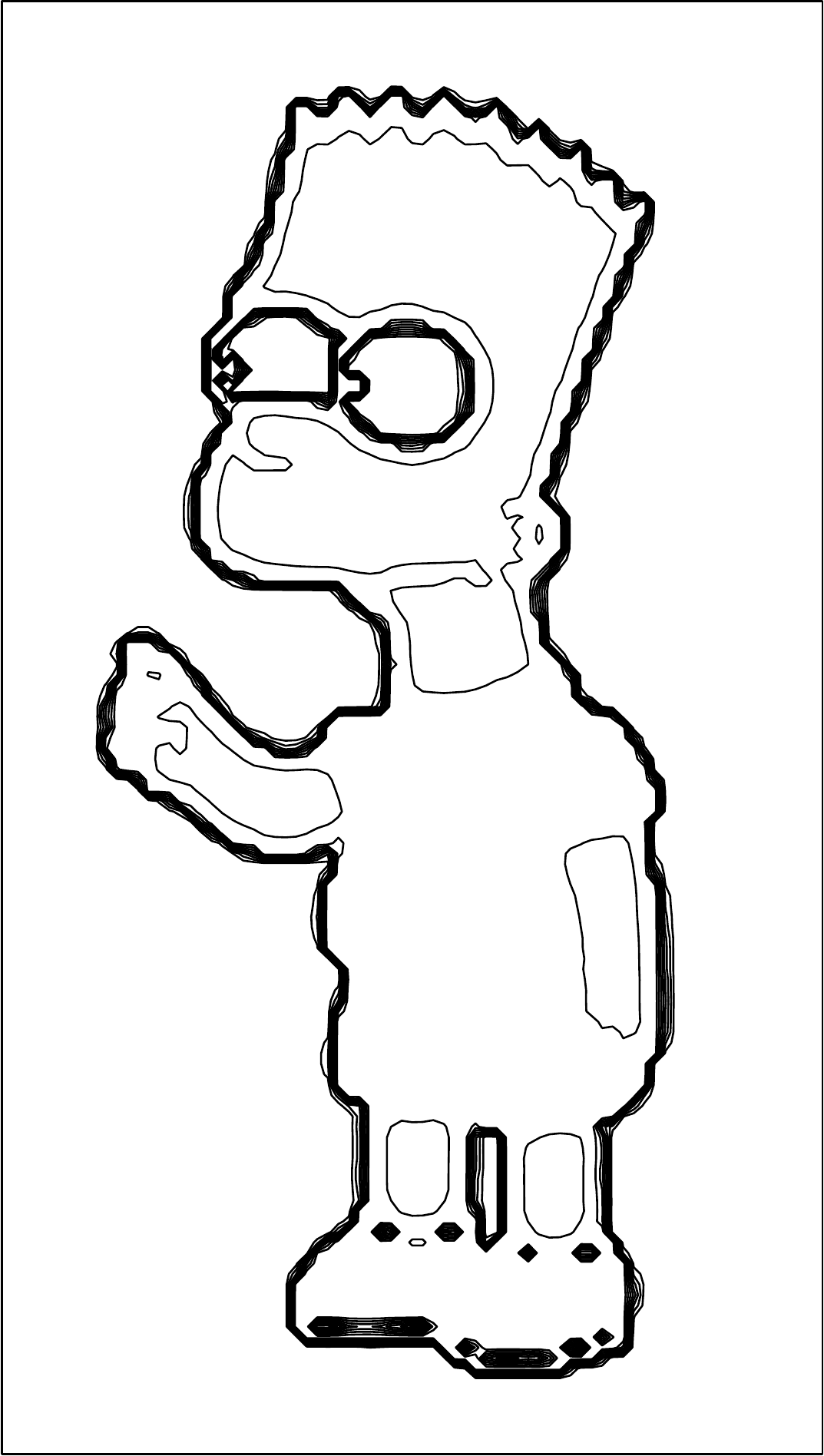}&
\vspace{-4.25cm}\includegraphics[height=.175\textwidth, width=4cm, angle=-90]{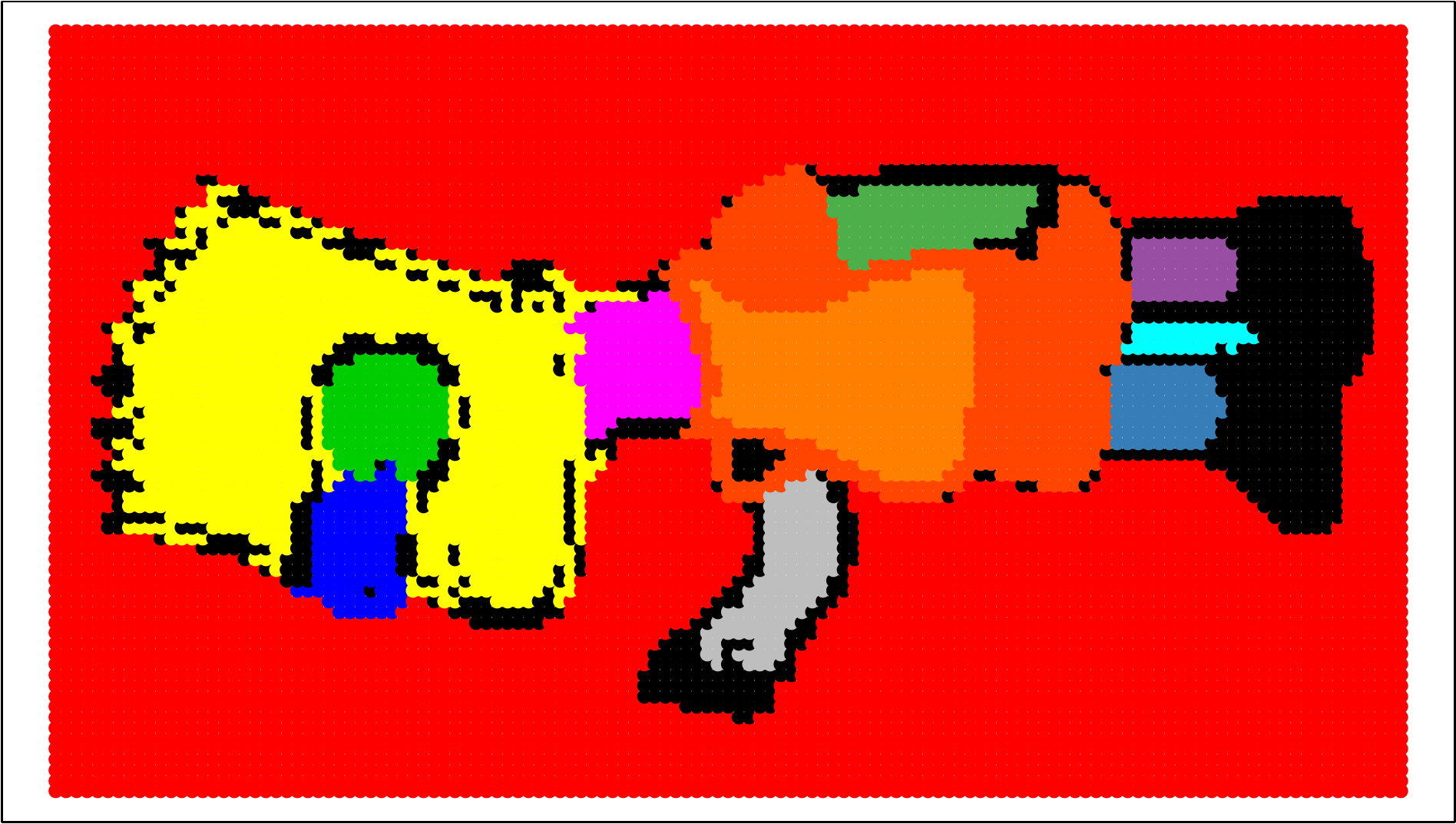}&
\vspace{-4.25cm}\includegraphics[height=.175\textwidth, width=4cm, angle=-90]{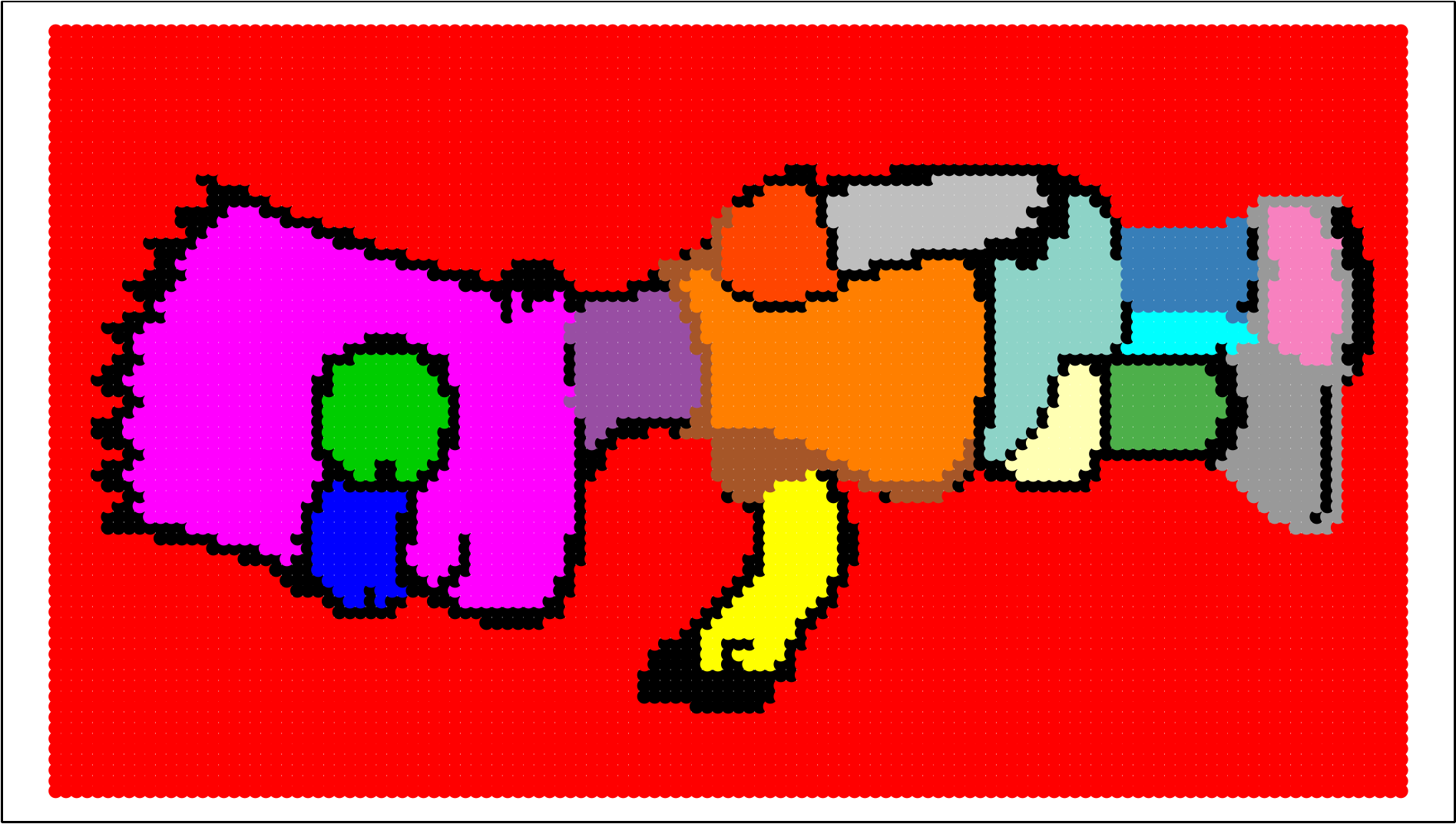}
\end{tabular}
\end{center}
\caption{Segmentation results. Segments have been assigned arbitrary colors, except for the thresholding segmentation, where segments are either black or white by construction.}\label{fig:bart} 
\end{figure*}

\begin{figure*}[tb]
\begin{center}
\begin{tabular}{p{3.1cm}p{3.1cm}p{3.1cm}p{3.1cm}p{3.1cm}}
\fontsize{7}{10}\selectfont {original image} & \fontsize{7}{10}\selectfont {$K-$means segmentation, $K = 5$} & \fontsize{7}{10}\selectfont {thresholding segmentation} & \fontsize{7}{10}\selectfont {Sobel segmentation} & \fontsize{7}{10}\selectfont {nonparametric segmentation, $h = h_N$}\\
\vspace{-0cm}\includegraphics[width=.175\textwidth,height=3cm]{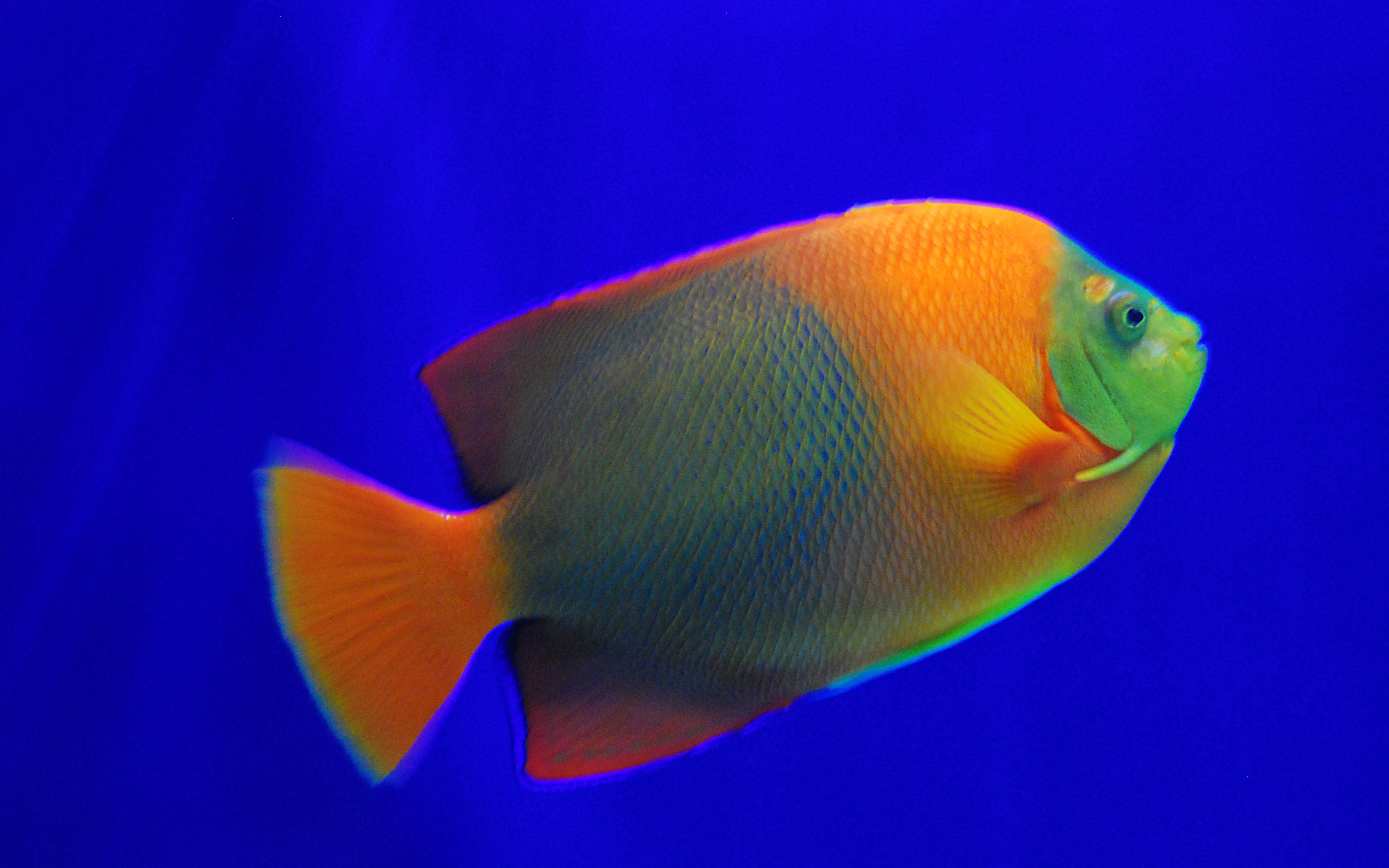}&
\includegraphics[height=.18\textwidth, width=3cm,angle=-90]{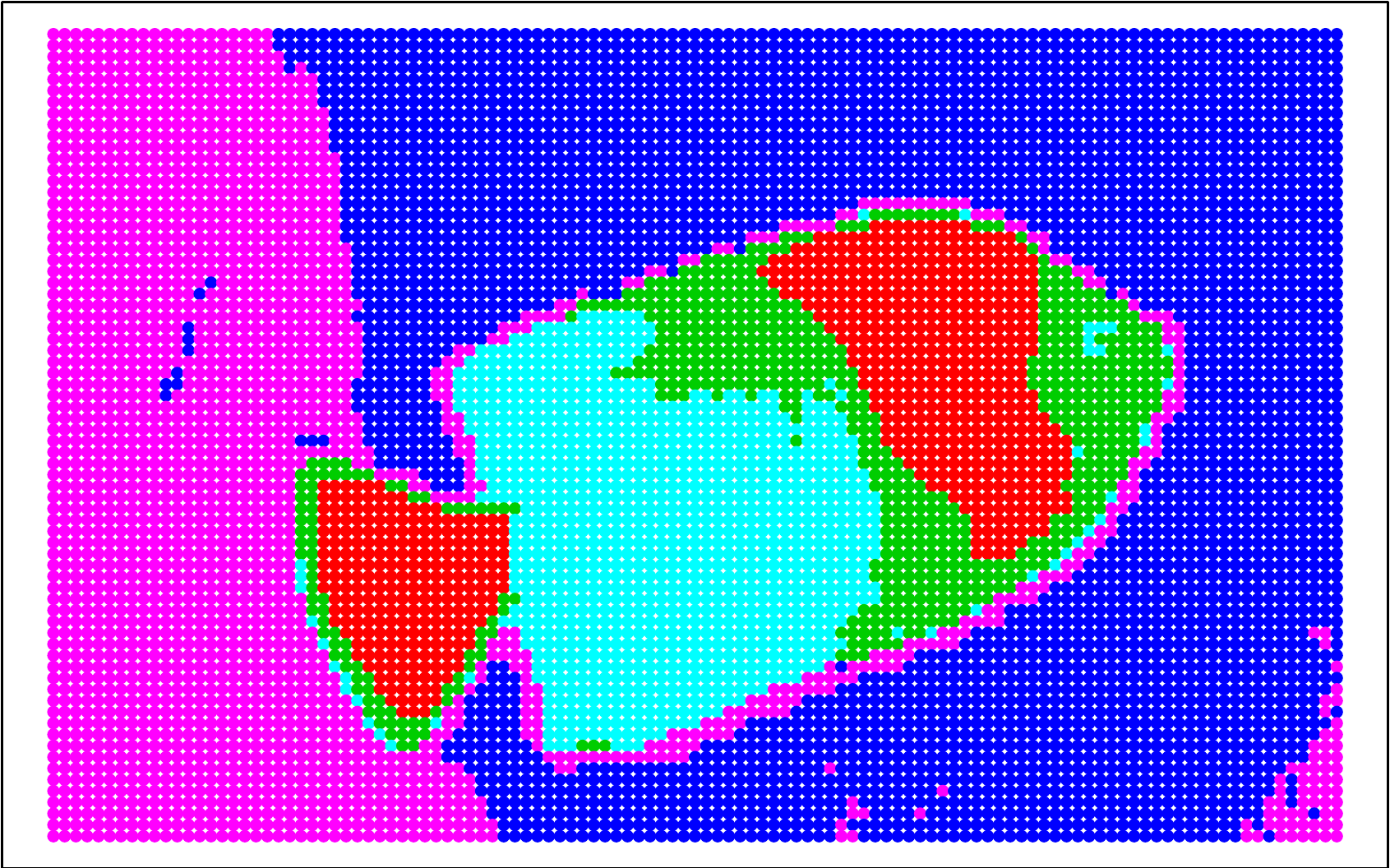}&
\includegraphics[width=.175\textwidth, height= 3cm,angle= -180]{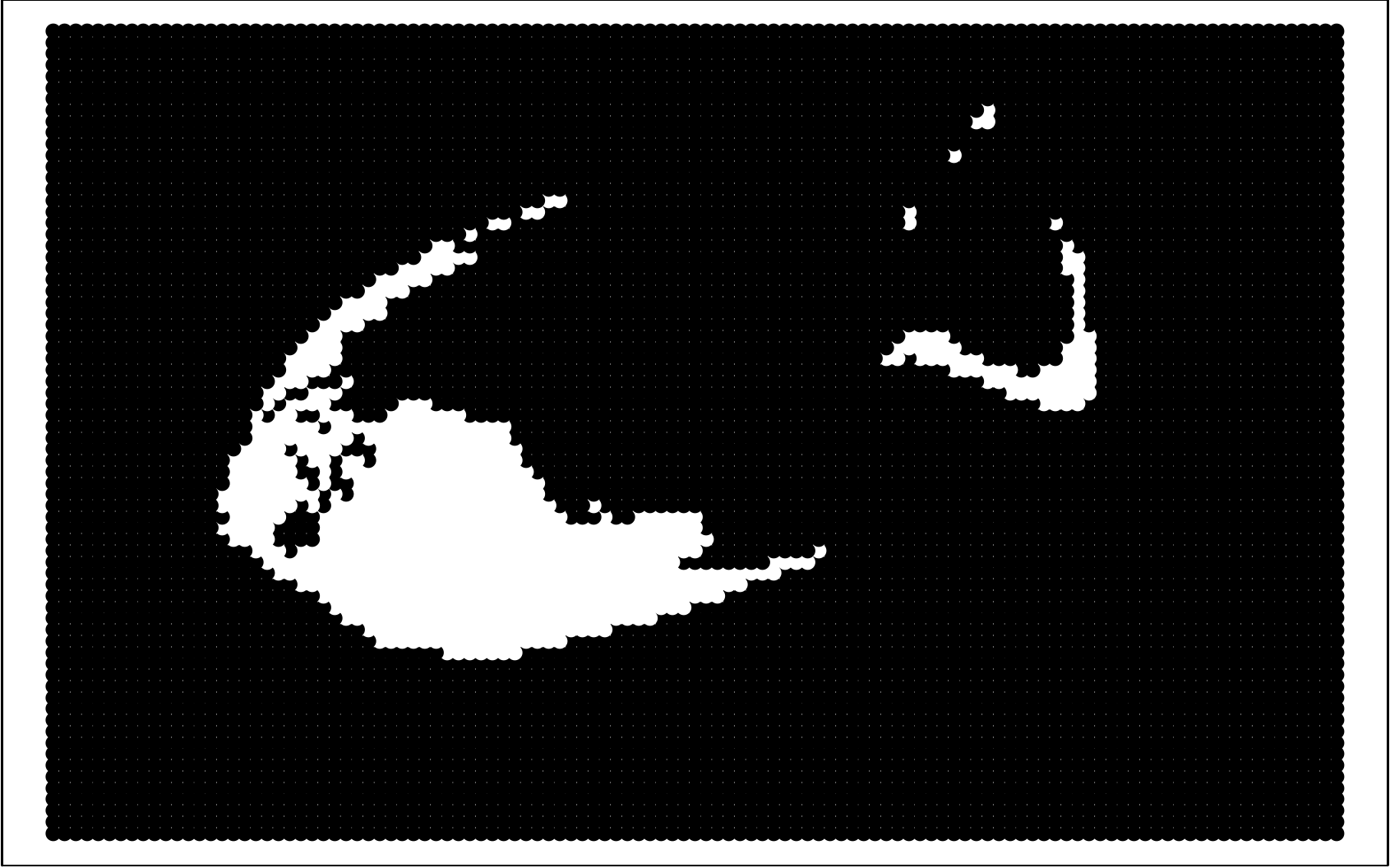}&
\raisebox{-\height}{\includegraphics[width=.175\textwidth, height=3cm]{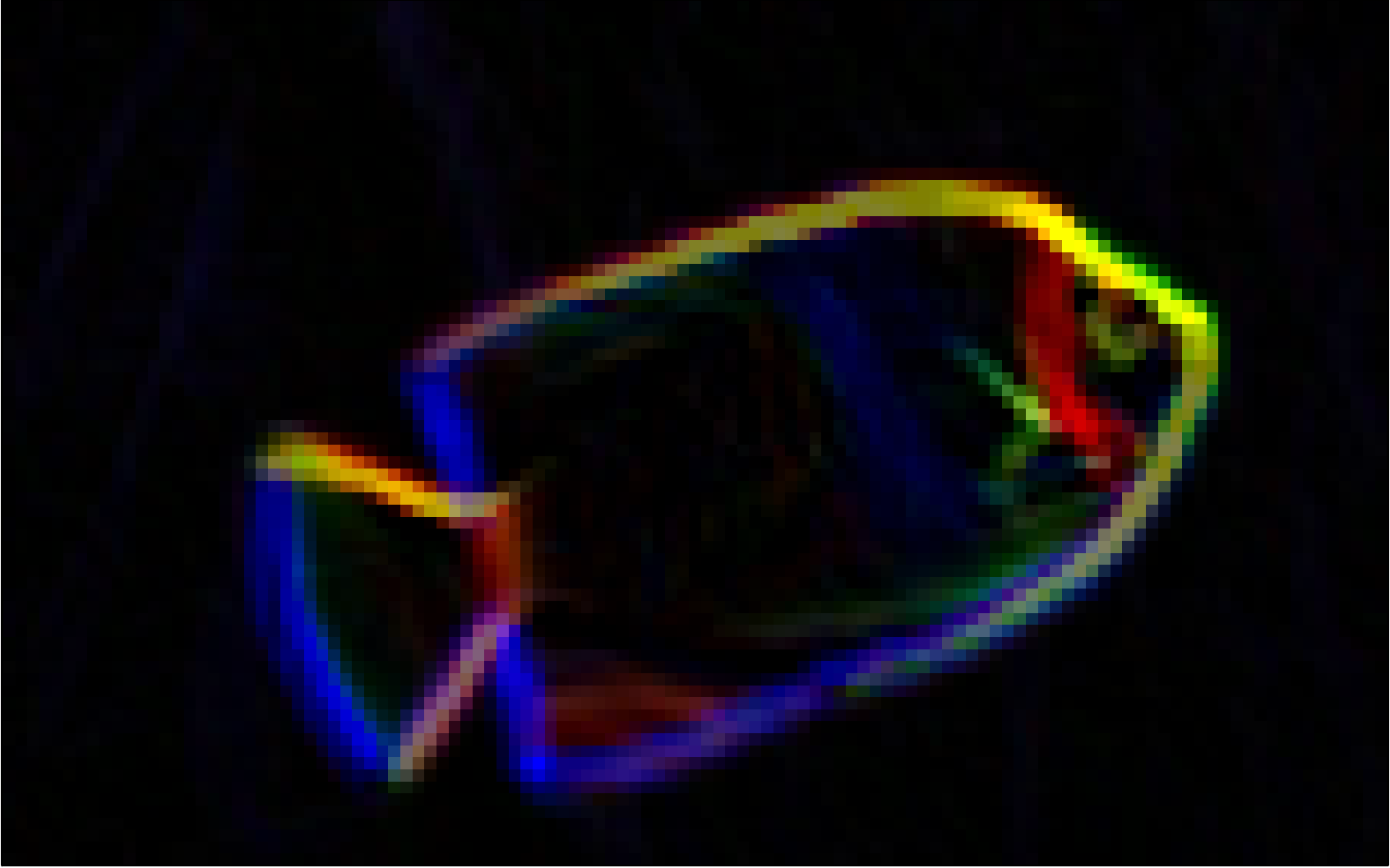}}&
\includegraphics[height=.175\textwidth, width=3cm,angle=-90]{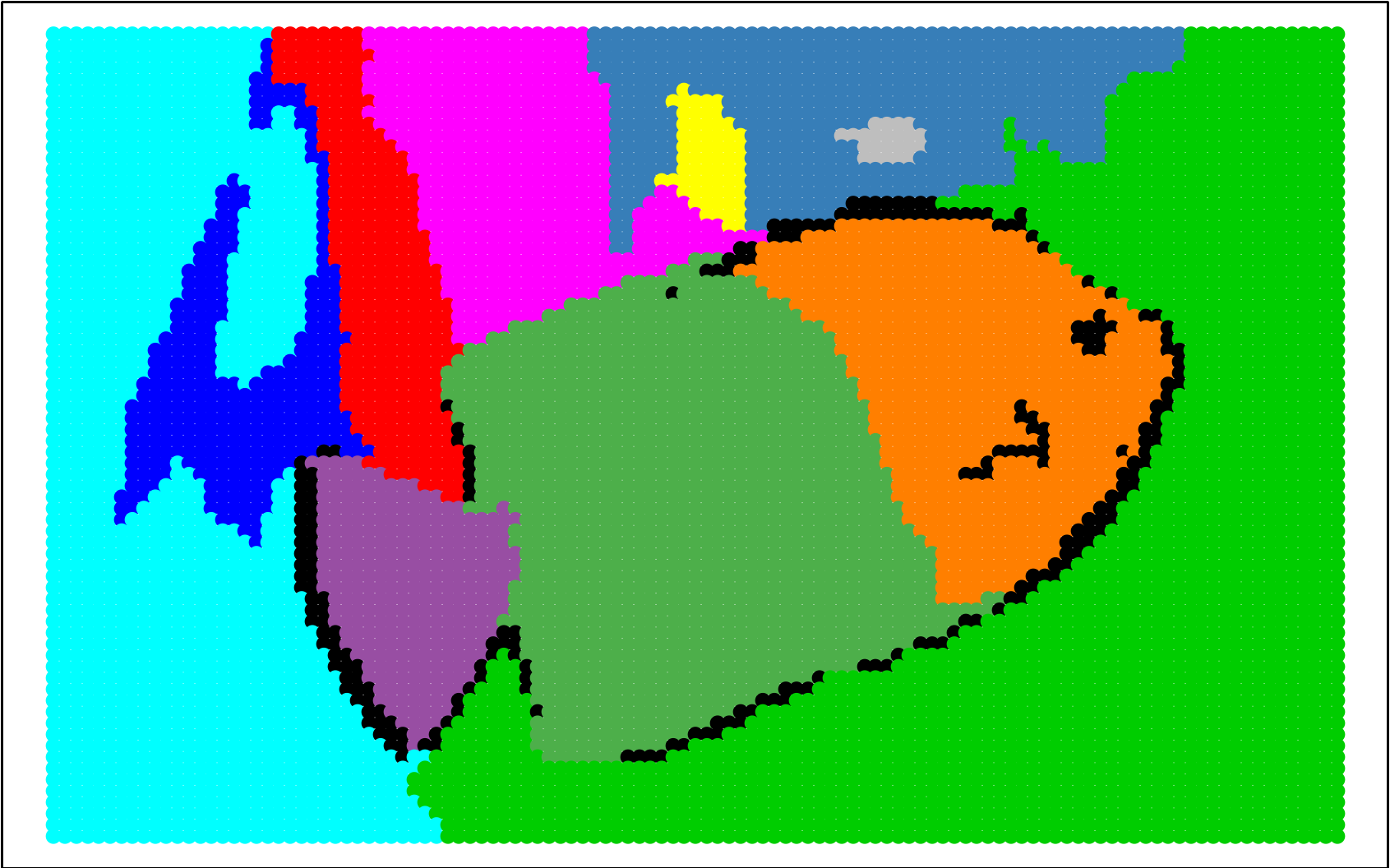}\\
\fontsize{7}{10}\selectfont { segment cores, $h = h_N$} & \fontsize{7}{10}\selectfont { cluster tree, $h = h_N$} & \fontsize{7}{10}\selectfont { density contours, $h = h_N$} & \fontsize{7}{10}\selectfont { nonparametric segmentation, $h = 0.75h_N$} & \fontsize{7}{10}\selectfont { nonparametric segmentation, $h = 1.25h_N$}\\
&&&&\\
\vspace{-3.35cm}\includegraphics[height=.175\textwidth, width=3cm, angle=-90]{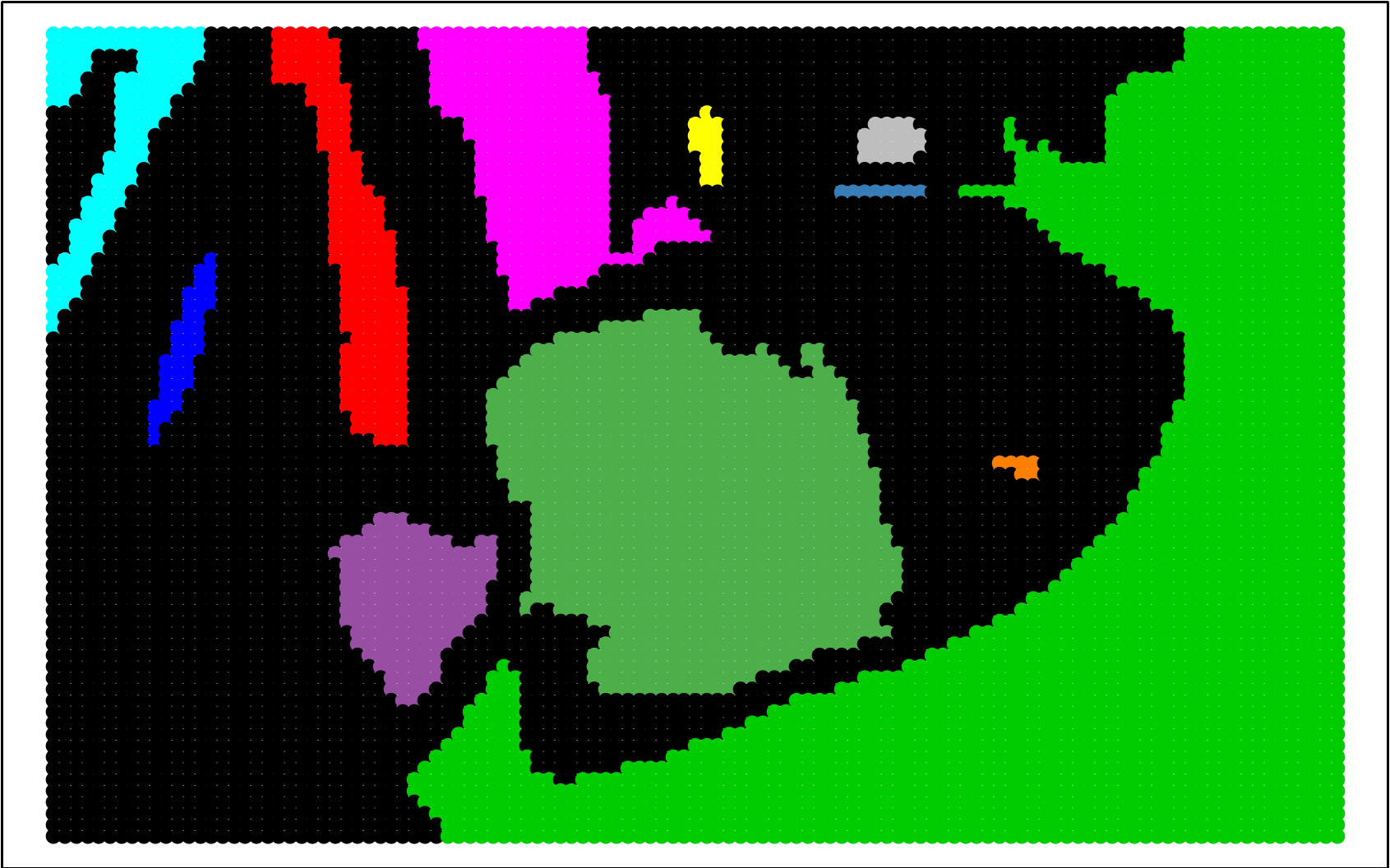}
&
\includegraphics[height=3cm, width=.175\textwidth]{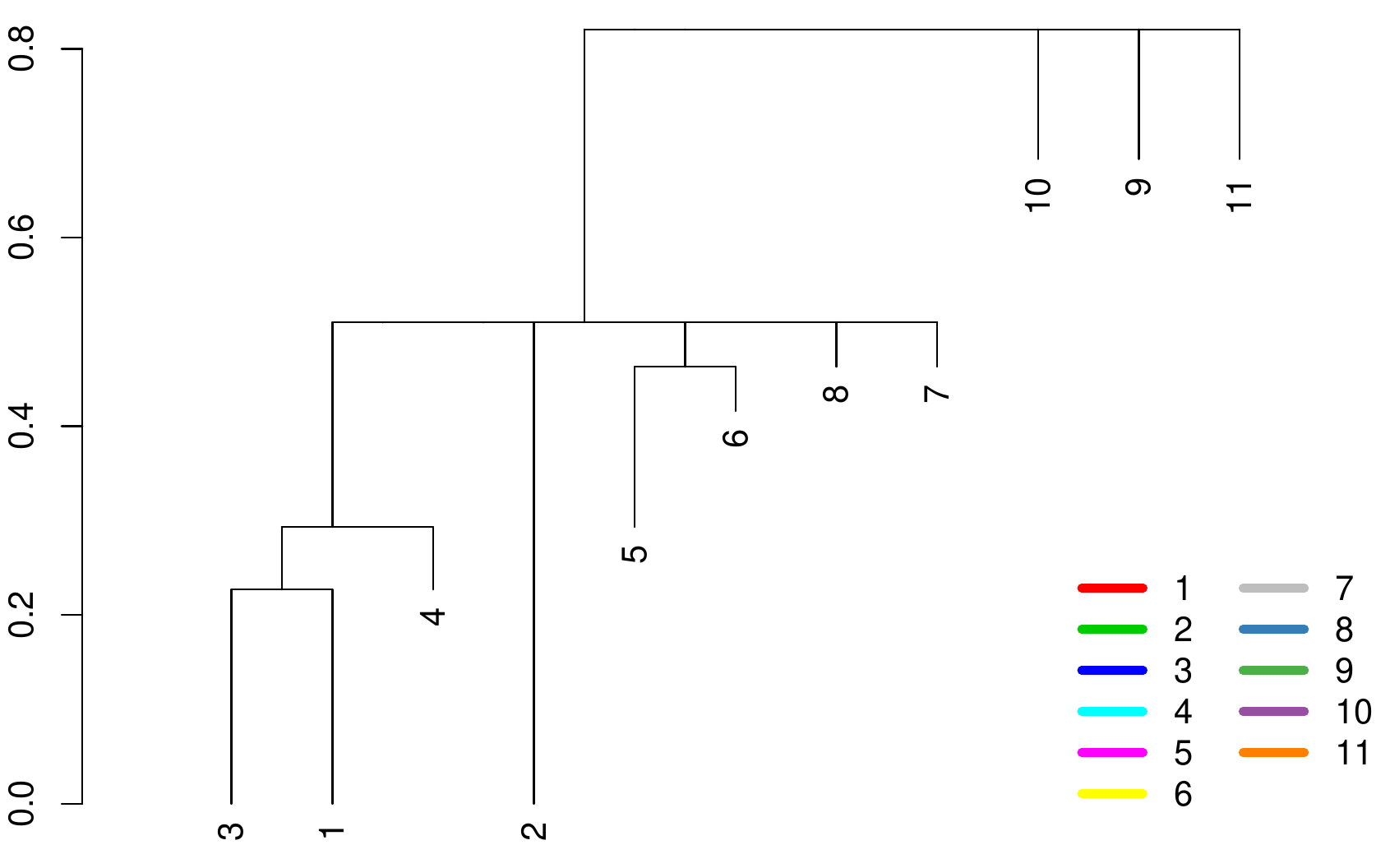}&
\includegraphics[height=3cm, width=.175\textwidth]{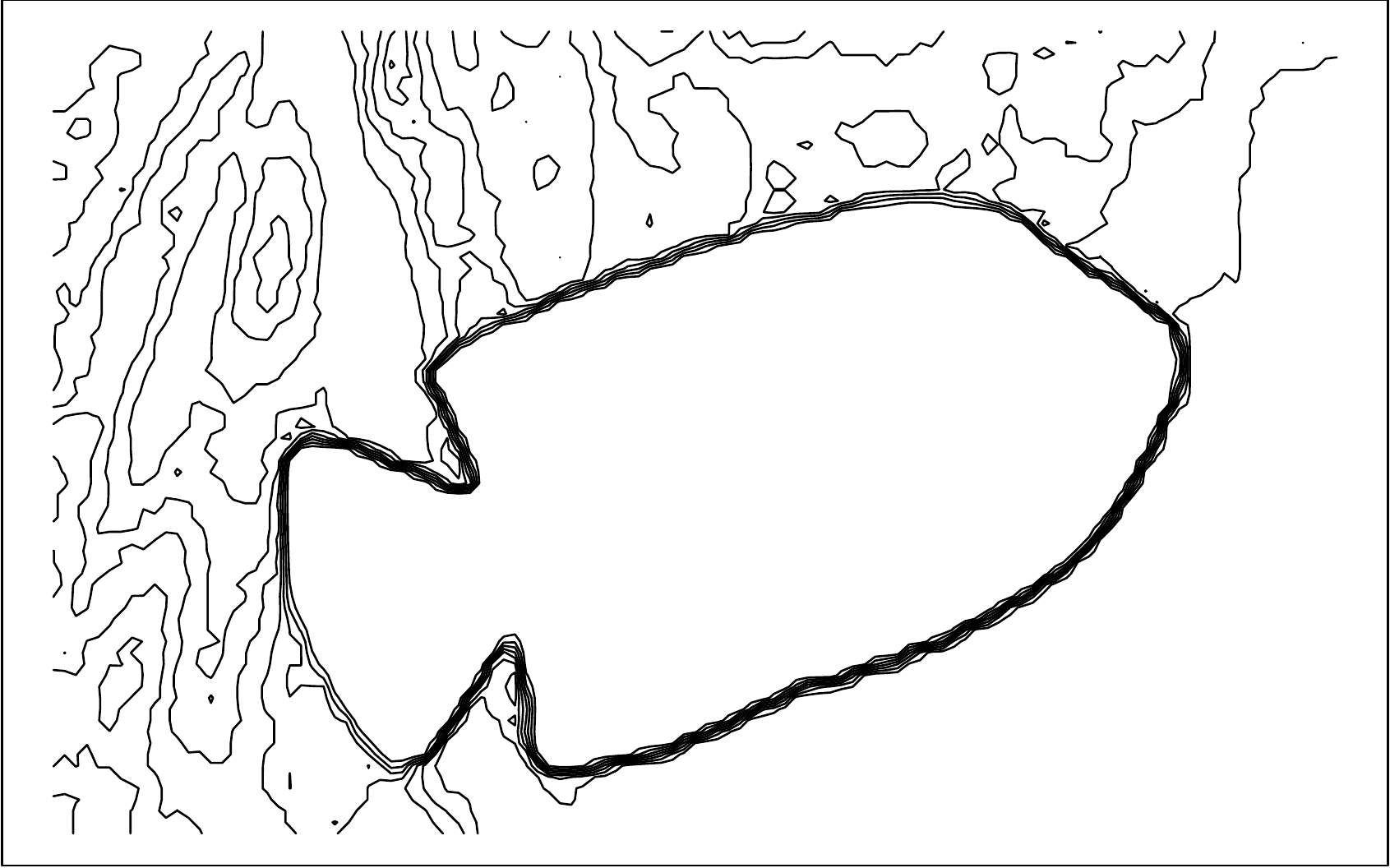}&
\vspace{-3.35cm}\includegraphics[height=.175\textwidth, width=3cm, angle=-90]{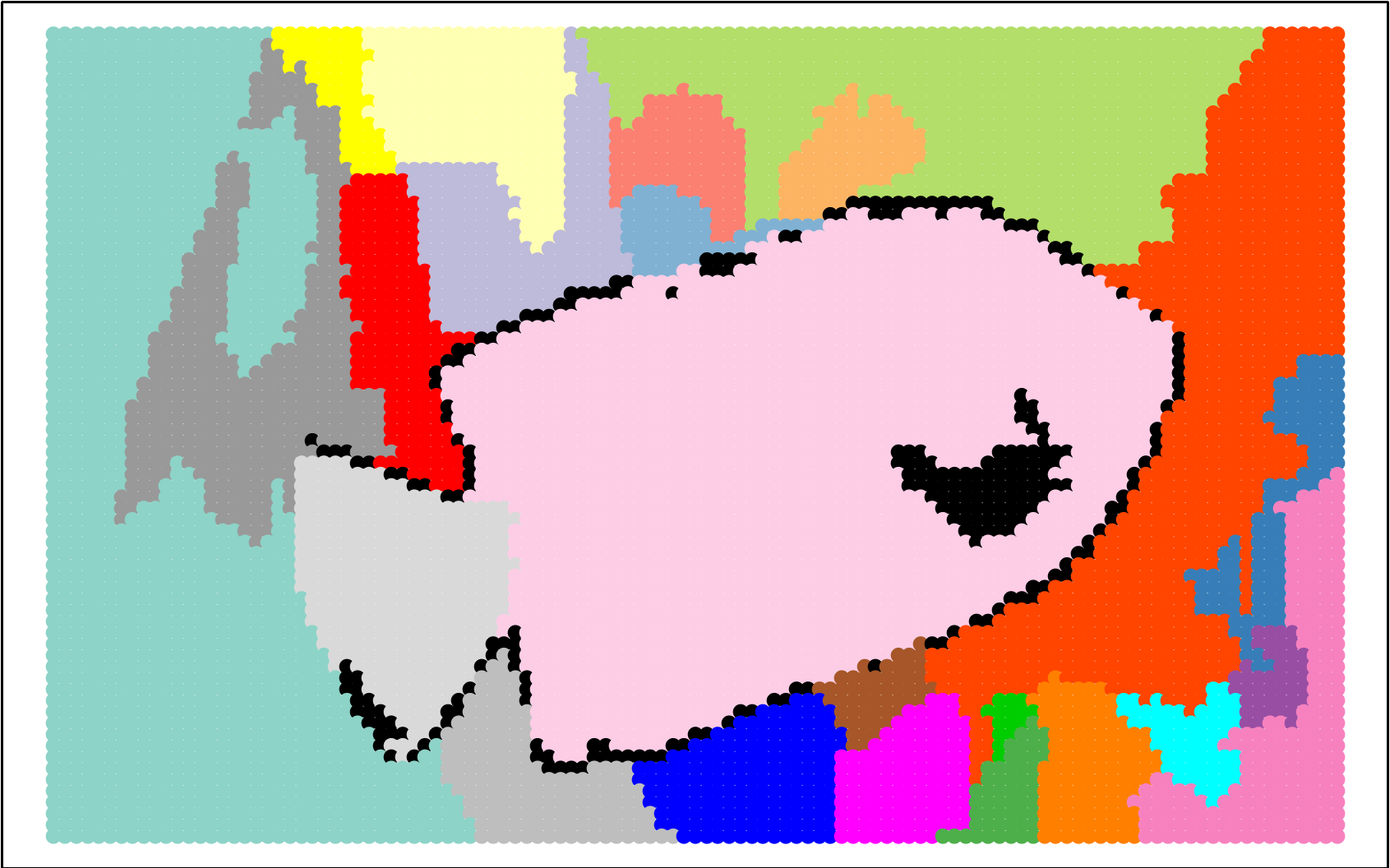}
&
\vspace{-3.35cm}\includegraphics[height=.175\textwidth, width=3cm, angle=-90]{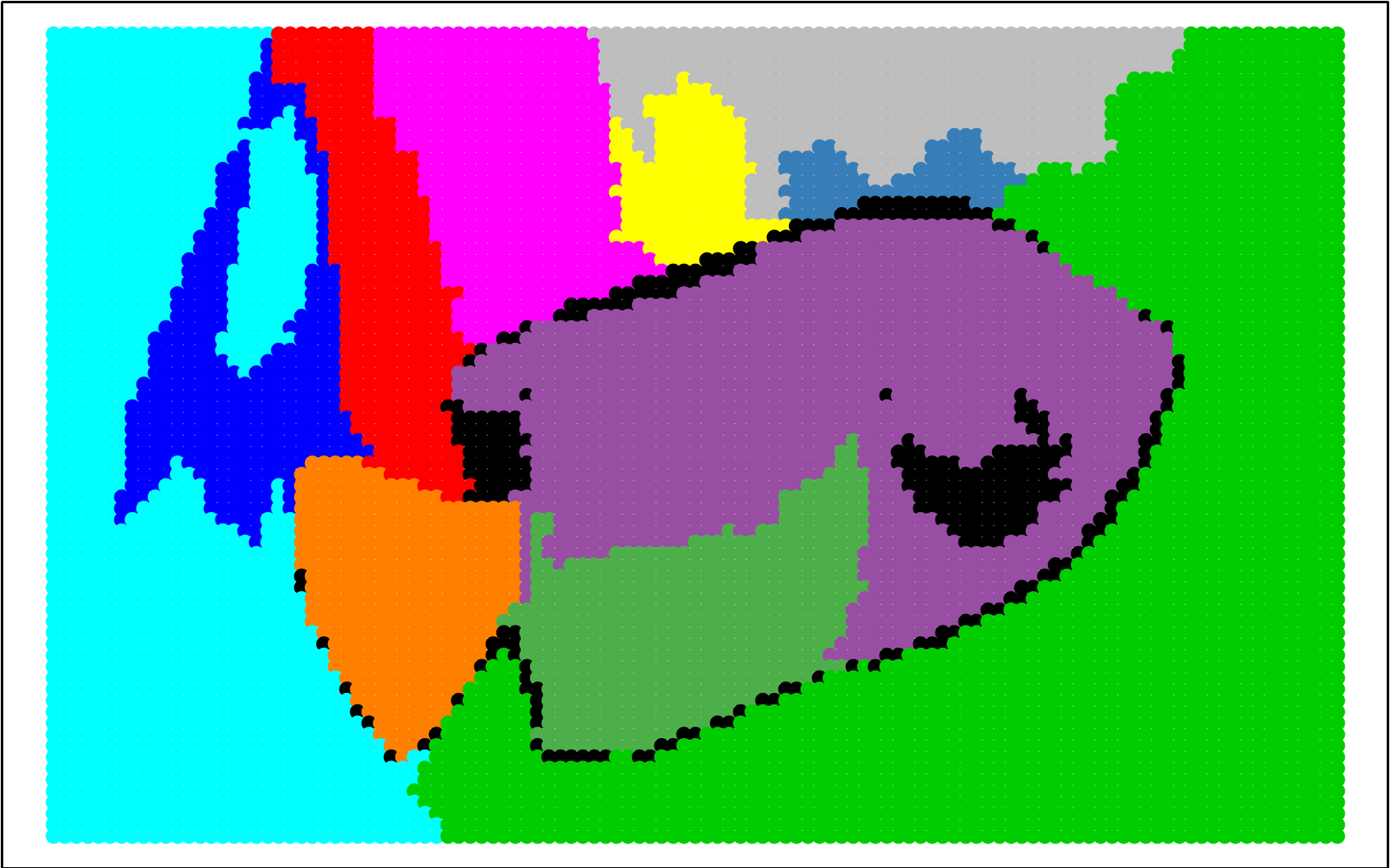}
\end{tabular}
\end{center}
\caption{Segmentation results. Segments have been assigned arbitrary colors, except for the thresholding segmentation, where segments are either black or white by construction.}\label{fig:fish} 
\end{figure*}

The top left panels of Figures \ref{fig:bnsquare}, \ref{fig:ein} display two examples of grayscale images, selected for the analysis due to their different characteristics and
different degrees of difficulty in segmentation. Multichannel images are displayed in the top left panels of Figures \ref{fig:bart} and \ref{fig:fish} . 

The benchmark procedures work satisfactorily, in grayscale and multichannel images. Thresholding is intrinsically limited, as it identifies two segments only by construction.
It is, nevertheless, able to reconstruct the broad features of the image, even when it is particularly challenging. 
See, \emph{e.g.} the famous Einstein's grimace, in Figure \ref{fig:ein}, not easy at all to be segmented, yet very well recognizable even in the binary reconstruction via thresholding. 
A self-evident limitation occurs when similar color shades characterize adjacent segments, since the dichotomous choice between black or white segments unavoidably determine a corrupt reconstruction of the image. See Figures \ref{fig:bnsquare} as an example of this behaviour. 
 
Despite its simplicity, $K-$means behave very well, being able to reconstruct accurately most of the images where the segment distinction is unarguable (Figures  \ref{fig:bnsquare}, \ref{fig:bart}). 
On the other hand, the procedure requires the number of segments to be known in advance, and a remarkable head start has been then granted by providing the true number of segments as an input. 
In some applications such number might be known a priori; consider, for example, X-ray or magnetic resonance medical images, where the number of segments can be set based on anatomy knowledge. However, if the purpose of image segmentation would be to attempt a first automatization of the diagnostic process, setting the number of segments to its `normal' value, would prevent detection of fractures or anomalies, thus going in the opposite direction.  
A further feature of $K-$means is its complete noncontextuality; since distances from the cluster centroids are computed on the basis of the color only, disconnected segments might be assigned to the same cluster just for sharing a similar color, disregarding the adjacency structure. Depending on applications, this characteristic may be sensible or not. For instance, in Figure \ref{fig:bart}, assigning head and body to the same segment is particularly consistent. On the other hand, a complete noncontextuality may lead to the identification of meaningless clusters, as it happens for Figure \ref{fig:ein} whose segmentation by $K-$means results in a pixelated image. 

The Sobel algorithm recognizes the edges between different segments generally in a very detailed way. This is particularly evident not only in simple examples with neat contours, like Bart Simpson (Figure \ref{fig:bart}) but also in challenging images like the fish (Figure \ref{fig:fish}). On the other hand, detection of the edges can be even too much detailed (Figure \ref{fig:ein}), with a risk of not identifying the salient regions of the image. Similarly to thresholding, it fails when there is small change in the color shades between adjacent segments, as it occurs with Figure \ref{fig:bnsquare}. Furthermore, the edges do not form closed boundaries, thus possibly requiring the possible application of some further algorithm to complete the scope and define the segments.

The performance of the proposed nonparametric procedure are generally satisfactory when applied to both multichannel and grayscale images. The procedure does not result challenged by the need of distinguishing contours of assorted segments, both for size and nonconvex shapes, as especially evidenced by Figures \ref{fig:ein} and \ref{fig:bart}. Consistently with the results of the simulation study, it 
is especially able to identify segments as connected regions characterized by uniformity of color, but performs well also when applied to image featured by shaded colors.  
On the con's side, the procedure is somewhat sensitive to perceive color differences even when they are not distinguishable with the unaided eye at once, thus resulting in oversegmenting the image. 
As the method mainly hinges on the density, which is built on the image colors, it is in principle framed within the class of the noncontextual algorithms. However, it takes in some information about the spatial relationship between the pixels since each segment is, by construction, a (high density) connected set which is disconnected from the other segments. This characteristic prevents pixeled segmentations like $K-$means and, on the other hand does not allow the identification of unique segments as disconnected regions sharing the same color (see Figure \ref{fig:bart}, where Bart's body and head are classified as different segments). 

Note that most of the segmented images include a number of black-colored regions, corresponding to unallocated pixels, and typically located at the borders of the segments. These are associated to low-density areas evaluated in the second step of the segmentation procedure. As discussed at the end of Section \ref{sec:procedure}, in those cases, none of the pixels already assigned to the segment presenting maximum density (\ref{eq:unallocated}) is adjacent to the ones under evaluation, \emph{i.e.}, their color is not similar to the color of any other adjacent segment. In the analysis, these pixels are left unallocated, to enlighten the low degree of confidence in their classification. 
Related to this aspect, and depending on subject-matter considerations, the procedure allows the opportunity of not allocating pixels that are not belonging to the cluster cores. The values of $\hat f_m$ in \eqref{eq:unallocated}, suitably normalized, provide a degree of confidence in the allocation, in the guise of fuzzy clustering schemes. This is especially useful in all the images where colors are homogeneous and segments well separated, as unallocated pixels mostly identify the boundaries of the segments (first bottom panel of Figures \ref{fig:bnsquare} to \ref{fig:fish}).   

With regard to the cluster tree (second bottom panel of Figures \ref{fig:bnsquare} to \ref{fig:fish}), it works effectively with multichannel images in establishing a hierarchy of cluster-merging which can be associated to different levels of resolution. In general, when scanning the density for varying $\lambda$, two segments appear disconnected because they are separated by some pixels having lower density color. At a lower level of $\lambda$, also these latter pixels have density above the threshold and merging can occur due to the spatial connectedness of all the involved pixels. In the Bart image, for example, clusters that are kept separated due to small color differences (see the face and the neck) are the first to be aggregated through the cluster tree. Similarly, in the fish image (Figure \ref{fig:fish}), at some high density level the different segments composing the sea are aggregated. 
 {In grayscale images similar conclusions may be drawn. In the Einstein image, for instance, highest-density aggregations of the tree branches entail merging the segments of the background. Yet, establishing  
a meaningful hierarchy of the segments is less easy with grayscale images, where also the human eye is challenged to aggregate clusters without resorting to subject-matter considerations and on the basis of the color only. }

In general, the density function results in an effective tools to identify the main features of the images, and density contours work well as edge detectors of the segments when $h = h_N$. See the third bottom panel of Figures \ref{fig:bnsquare} to \ref{fig:fish}.  Additionally, the segmentation is quite robust to variation of $h$. This is especially true for simple images with neat contours as the Greyscale raws and Bart Simpson (IV and V bottom panels of Figures \ref{fig:bnsquare} and \ref{fig:bart}), where comparable segmentations are produced over the range of considered values of $h$. As expected, more challenging images tend to be oversegmented for small $h$ as seen, for instance, in the background of the Einstein and fish images (IV bottom panel of Figures \ref{fig:ein} and \ref{fig:fish}). A large value of $h$, on the other hand, smoothes the density and results in segment aggregation, as seen in the last bottom panel of Figures \ref{fig:ein} and \ref{fig:fish}. This can be seen as a way to reduce the risk of over-segmentation. 
\section{Concluding remarks}

Image segmentation is a complicated task whose implementation cannot, in general, leave aside subject-matter considerations. 
All this considered, the proposed procedure is framed halfway between contextual and noncontextual segmentation algorithms, and may be then applied to a variety of situations. 
It can be either applied fully automatically, or be richly customized, depending on the goals of the segmentation. It is provided with some useful tools that may integrate the output of segmentation, as an estimate of the density of the pixels, which may be used to determine the degree of confidence about the segment allocation, as well as for edge detection, and the cluster tree, which allows for displaying different levels of resolution of the segmentation itself. 

\section*{Supporting information}

Some further examples are available in the Supplementary Material, as part of the online article.

\section*{Acknowledgments}
I wish to thank Fabio Barbaro, who first approached the embrionic ideas of this work when drafting his degree thesis.
\bibliography{biblio}
\end{document}